%% file: main.tex
\documentclass{article}
\usepackage{iclr2026_conference,times}

\input{math_commands.tex}

\usepackage{hyperref}
\usepackage{xurl} 
\usepackage{graphicx}
\usepackage{booktabs} 
\usepackage{enumitem} 
\usepackage{makecell} 
\usepackage{array}
\usepackage[table]{xcolor}
\usepackage{multirow}
\usepackage[linesnumbered, ruled, vlined]{algorithm2e}

\SetCommentSty{mycommfont}
\usepackage{xspace} 
\newcommand{\mask}{[\texttt{MASK}]\xspace}
\usepackage{subcaption}
\usepackage{wrapfig}
\usepackage[dvipsnames]{xcolor}

\newcommand{\gain}[1]{\textit{\color{Green}#1}}

\title{Soft-Masked Diffusion Language Models}

\author{Michael Hersche$^{1}$, Samuel Moor-Smith$^{1,2}\thanks{Research conducted at IBM Research -- Zurich.}$\,\,, Thomas Hofmann$^{2}$, Abbas Rahimi$^{1}$\\
\And
\normalfont $^{1}$IBM Research -- Zurich, $^{2}$Department of Computer Science, ETH Zürich}

\iclrfinalcopy

\begin{document}

\maketitle

\begin{abstract}
Diffusion models have demonstrated strong potential in language modeling, offering various advantages over traditional autoregressive approaches. 
Their ability to generate and revise entire responses in parallel enables faster generation and built-in self-correction mechanisms.
Most modern diffusion-based language models employ masked diffusion, where decoding involves iteratively processing masked tokens based on a binary decision: either retaining the mask or replacing it with the predicted token. 
However, this binary choice discards valuable predictive information when the mask is retained.
To address this limitation, we introduce \textit{soft-masking (SM)}, a novel method that dynamically blends the embedding of the mask token with the embeddings of the top-$k$ predicted tokens from the previous decoding step, for each retained mask.
This provides the model with a more informative prior, preserving context from earlier computations and allowing partial information about masked tokens to propagate beyond a single step. 
We propose a training methodology that efficiently adapts masked diffusion language models to incorporate SM.  
We demonstrate that training a 169M parameter model from scratch with SM yields superior perplexity and MAUVE scores compared to binary masking baselines. Similarly, a pretrained model can be enhanced with SM through continued pretraining.
Finally, we finetune two state-of-the-art diffusion models, Dream-7B and Dream-Coder-7B, with SM.
SM consistently improves performance across multiple coding benchmarks, particularly in high-throughput settings.\footnote{Code at \url{https://github.com/IBM/soft-masked-diffusion-language-models}}
\end{abstract}

\begin{figure*}[h!]
\centering
\includegraphics[width=\textwidth]{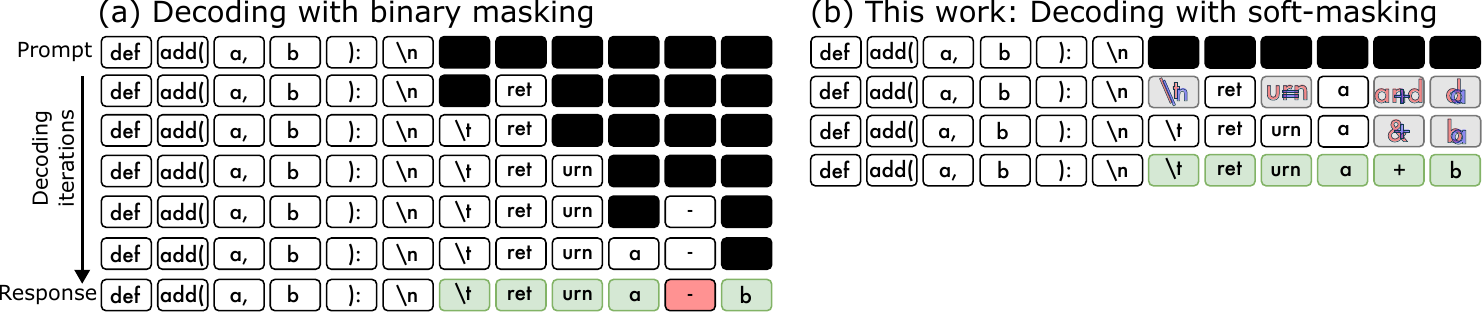}
\caption{
Illustrative answer generation using masked diffusion language models (MDLMs) via iterative decoding with (a) standard binary masking or (b) our proposed soft-masking.
Our soft-masking enriches the feedback for the next decoding step by superposing the masked tokens with the previously predicted top-$k$ candidates, enabling more accurate and faster generation.
}
\label{fig:concept}
\end{figure*}

\section{Introduction}

Generative large language models (LLMs) have transformed natural language processing. 
LLMs typically operate in an autoregressive (AR) mode~\citep{vaswani_attention_2017, brown_language_2020}, where the next token in a sequence is predicted based on the previously generated tokens. 
While AR models have proven highly effective, their sequential nature makes inference computationally expensive, leading to high latency and costs. 
These inference costs are particularly pronounced in large reasoning models~\citep{openai_learning_2024, deepseekai2025short, qwen_qwq_2024}, where different solutions are validated through sequential exploration via chain-of-thought (CoT)~\citep{wei_chainofthought_2022}.

As a potential remedy, recent work has shown that incorporating continuous feedback, rather than relying solely on discrete, sampled tokens, can improve AR model's performance~\citep{hao_training_2025, zhuang_mixture_2025, zhang_soft_02}.
Such continuous feedback encodes multiple potential solutions in superposition~\citep{zhu_reasoning_2025}, enabling simultaneous exploration of diverse paths and thereby potentially reducing the number of generated tokens.
However, training AR models with continuous feedback is slow, due to the sequential reliance on previous continuous token outputs.

As a promising alternative, diffusion models---originally developed for continuous domains in vision~\citep{sohl-dickstein_deep_2015, ho_denoising_2020, song_scorebased_2020}---have recently been adapted to natural language processing. These diffusion \emph{language} models (DLMs) offer key advantages over AR models, including accelerated sampling~\citep{inception_mercury_2025}, controllable generation~\citep{li_diffusionlm_2022}, bidirectional modeling, and built-in self-correction~\citep{ye_diffusion_2024}.  
Besides being more data-efficient than AR-models in training~\citep{ni_diffusion_2025, prabhudesai_diffusion_2025}, 
they are particularly beneficial in non-causal tasks such as coding~\citep{nie_large_2025, gong_diffucoder_2025, xie_dreamcoder_2025}.
They consist of a forward process, which gradually corrupts data, and a backward process, which iteratively reverses this corruption to generate coherent outputs.

Masked DLMs (MDLMs) have emerged as the most scalable and effective approach. 
They implement the forward process as a categorical transition function, mapping tokens to an absorption state, typically represented by a \mask token~\citep{austin_structured_2021, campbell_continuous_2022, lou_discrete_2024, sahoo_simple_2024, ou_your_2025, shi_simplified_2024}.
During decoding, the model makes a binary choice for each mask token: either replace it with a predicted token or retain the \mask (see \Figref{fig:concept}a). 
This discrete formulation allows for improved training and has enabled the development of large-scale MDLMs across both open-source~\citep{gong_scaling_2025, nie_large_2025, ye_dream_2025, xie_dreamcoder_2025} and commercial~\citep{inception_mercury_2025, deepmind_gemini_2025} initiatives.

Despite their scalability, MDLMs are fundamentally constrained by the binary unmasking process, which discards valuable predictive information. 
Likewise, AR models commit to a single discrete sampling decision, with no opportunity for refinement. 
Motivated by the success of continuous feedback mechanisms in AR models, which preserve and leverage uncertainty over multiple candidates, we propose a new feedback mechanism for MDLMs that propagates this rich predictive information throughout the generation process.

\paragraph{This work: Continuous feedback in MDLMs via soft-masking}
We introduce soft-masking~(SM), a simple yet effective mechanism for incorporating continuous feedback into MDLMs, as illustrated in~\Figref{fig:concept}b.
During decoding, SM enriches the \mask state with a convex combination of the top-$k$ predicted tokens, weighted by their confidence scores.
This allows the model to retain and propagate partial information across decoding steps, rather than discarding it through binary masking decisions.
Our method integrates seamlessly into existing MDLM architectures, requiring minimal adaptation.
Our contributions are as follows: 
\begin{itemize}[itemsep=-0.0em, topsep=-0.1em]
\item We propose soft-masking (SM), a novel decoding mechanism that enhances the expressiveness of the \mask token in MDLMs. SM only adds three additional parameters, which can be efficiently learned together with the MDLM parameters using a parallelizable training procedure that enables MDLMs to leverage the richer feedback.
\item We show that a 169M-parameter MDLM with SM trained on OpenWebText can enhance both validation perplexity and MAUVE scores in unconstrained generation. Further, SM can be integrated into a pretrained MDLM via pretraining continuation for only 100k steps. 
\item We demonstrate that SM generalizes to large-scale MDLMs by applying it to Dream-7B~\citep{ye_dream_2025} and Dream-Coder-7B~\citep{xie_dreamcoder_2025}. 
After minimal finetuning, SM improves accuracy on HumanEval and MBPP code generation benchmarks (including plus versions), particularly in high-throughput regimes with limited decoding iterations.
\item SM can be readily integrated into other MDLM efficiency enhancement techniques, such as unmasking and caching. We show that SM complements an advanced unmasking scheduler, ReMDM~\citep{wang_remasking_2025}, further improving unconstrained text generation quality. 
Moreover, SM can leverage caching and confidence-aware blockwise decoding from Fast-dLLM~\citep{wu_fastdllm_2026}, particularly improving generations in high-throughput settings.  
\end{itemize}

\section{Background: Masked Diffusion Language Models}
We begin by formulating general diffusion models~\citep{sohl-dickstein_deep_2015} and continue with the forward and backward diffusion processes in MDLMs~\citep{gong_scaling_2025}. 
We denote scalars with lower-case letters ($x$), vectors with bold lower-case letters ($\mathbf{x}$), sequences of length $T$ with a colon (e.g., $\mathbf{x}_{1:T})$, matrices with bold capital letters ($\mathbf{X}$), and the transpose with $\mathbf{X}^\top$. 

Diffusion models describe a \emph{forward} diffusion process as a Markov chain that progressively corrupts the original data: $q(\mathbf{x}_{1:T} | \mathbf{x}_0) = \prod_{t=1}^T q(\mathbf{x}_t | \mathbf{x}_{t-1})$. 
Here, $\mathbf{x}_0\sim q_{\text{data}}(\mathbf{x}_0)$ is drawn from the data distribution and $q(\mathbf{x}_t | \mathbf{x}_{t-1})$ describes the transition probability at step $t$. 
The marginalized target distribution ($q(\mathbf{x}_T)$) should be stationary and cheap to generate (e.g., a Gaussian distribution). 
A \emph{reverse} diffusion process aims to reconstruct the original data with a parameterized function $p_\theta(\mathbf{x}_{0:T})=p_\theta(\mathbf{x}_{T})\prod_{t=1}^T p_\theta(\mathbf{x}_{t-1} | \mathbf{x}_t)$. 

\subsection{MDLM Modeling}

\paragraph{Forward corruption}

We first focus on the corruption process for a single token; the extension to sequences is discussed later.
We represent language tokens as one-hot vectors $\mathbf{x} \in \{0,1\}^{|\mathcal{V}|}$, where $|\mathcal{V}|$ represents the cardinality of the vocabulary. 
The transition function in MDLMs is defined such that, at each step, the token either remains unchanged or is mapped to a designated absorption state: $\mask \in \mathcal{V}$.
The transition can be expressed as 
$
    q(\mathbf{x}_t|\mathbf{x}_{t-1}) = \text{Cat}(\mathbf{x}_t; \mathbf{Q}^\top_t\mathbf{x}_{t-1}),
$
where $\text{Cat}(\cdot,\mathbf{p})$ is the categorical distribution given a probability mass vector $\mathbf{p}\in \Delta^{|\mathcal{V}|-1}$, and $[\mathbf{Q}_t]_{i,j}$ denotes the transition probability from token $i$ to token $j$ at time $t$.
The marginal distribution after $s$ steps is: 
\begin{align*}
    q(\mathbf{x}_{s}| \mathbf{x}_0) = \text{Cat}(\mathbf{x}_s| \overline{\mathbf{Q}}^\top_s \mathbf{x}_0 ) = \alpha_s \mathbf{x}_0 + (1-\alpha_s) \mathbf{m},  
\end{align*}
where $\mathbf{m}$ is the mask token, $\overline{\mathbf{Q}}_s = \prod_{t=1}^s \mathbf{Q}_t$, and $\alpha_s$ describes the probability of retaining the original state. 
The schedule is typically chosen such that $\alpha_T = 0$, ensuring that the token is absorbed into the masking state with probability 1 at the final step $T$.
For example, a linear masking schedule with $\alpha_t =(1-t/T)$ is a popular choice~\citep{austin_structured_2021, gong_scaling_2025, nie_large_2025}. 

\paragraph{Reverse process}
Decoding aims to reverse the corruption process by iteratively denoising the data, starting from the absorbed (masked) state at time step $T$.
First, note that the forward transition probability between two time steps $0 \leq  s < t \leq T$ is given by:
\begin{align*}
    q(\mathbf{x}_s | \mathbf{x}_t) = \text{Cat}(\mathbf{x}_s; \overline{\mathbf{Q}}^\top_{t|s}\mathbf{x}_t), 
\end{align*}
where $\overline{\mathbf{Q}}_{t|s} = \overline{\mathbf{Q}}_s^{-1} \overline{\mathbf{Q}}_t$ represents the transition matrix from step $t$ back to step $s$.
Assuming access to the ground-truth token $\mathbf{x}_0$, the exact posterior transition from $\mathbf{x}_t$ to $\mathbf{x}_s$ can be computed via Bayes: 
\begin{align*}
    q(\mathbf{x}_s | \mathbf{x}_t, \mathbf{x}_0) &= \frac{q(\mathbf{x}_t | \mathbf{x}_s) q(\mathbf{x}_s | \mathbf{x}_0)}{q(\mathbf{x}_t | \mathbf{x}_0)} = 
\begin{cases}
\frac{\alpha_s - \alpha_t}{1 - \alpha_t} \mathbf{x}_0 + \frac{1 - \alpha_s}{1 - \alpha_t} \mathbf{m} & \text{if } \mathbf{x}_t = \mathbf{m}, \\
\mathbf{x}_0 & \text{if } \mathbf{x}_t \neq \mathbf{m}.
\end{cases}
\end{align*}
Since $\mathbf{x}_0$ is unknown during inference, we approximate the posterior using a learnable function $f_\theta(\mathbf{x}_t)$ that predicts the original token from the corrupted input $\hat{q}(\mathbf{x}_s | \mathbf{x}_t, \mathbf{x}_0) = p_\theta (\mathbf{x}_s | \mathbf{x}_t, f_\theta(\mathbf{x}_t))$. 
%
Here, a learnable function\footnote{While original denoising models use time conditioning ($f_\theta (\mathbf{x}_t, t)$),~\cite{ou_your_2025} present a method without time conditioning. We omit time conditioning in our theoretical formulation. However, we show experimentally that SM improves MDLMs \emph{with} (Section~\ref{sec:language_modeling}) and \emph{without} time conditioning (Section~\ref{sec:code_generation}).}
($f_\theta$) approximates the ground-truth ($\mathbf{x}_0$); hence, it imitates the denoising from step $t$ to step $0$. 
Substituting the approximation into the closed-form expression yields the parametric backward transition:
\begin{align}
  p_\theta(\mathbf{x}_s | \mathbf{x}_t) =   \frac{\alpha_s - \alpha_t}{1 - \alpha_t} f_\theta (\mathbf{x}_t) + \frac{1 - \alpha_s}{1 - \alpha_t} \mathbf{m}.  \label{eq:diff_decoding}
\end{align}

\paragraph{Reverse process in natural language processing}
The input consists of sequences of $L$ tokens: $\mathbf{x}^{1:L}_0$. 
Decoding begins from a fully masked sequence: $\mathbf{x}^{1:L}_T= \mathbf{m},...,\mathbf{m} $. 
At each time step $t$, the current sequence estimate is passed through a bidirectional model (e.g., a non-causal Transformer~\citep{vaswani_attention_2017, peebles_scalable_2023}), yielding token-wise probability distributions:
$
    \mathbf{p}^{1:L}_{t-1}= g_\theta(\mathbf{x}^{1:L}_t), \label{eq:pmf}
$
where $\mathbf{p}^l_{t-1}\in \Delta^{\mathcal{|V|}-1}$ is the predicted probability mass vector on the $|\mathcal{V}|$-dimensional simplex for token $l$.
Each probability mass vector ($\mathbf{p}^l_{t-1}$) is discretized using a sampling strategy (e.g., nucleus or argmax), yielding $\mathbf{\tilde{x}}^{1:L}_{t-1}$. 
Describing the sampling function with $h(\cdot)$, we can write the reverse process of the entire model as the functional composition of the sampling and model forward pass: $f_\theta= h \circ g_\theta  $.

\paragraph{Training objective}
Given a linear schedule of $\alpha_s$, the parameters ($\theta$) are optimized by minimizing: 
\begin{align}
    \Ls(\theta) =-\E_{t\sim U(0, 1),\mathbf{x}_0 \sim q_{\text{data}}(\cdot),\mathbf{x}_t \sim q(\cdot|\mathbf{x}_0}) \left[ \frac{1}{t} \sum_{i=1}^{L} \mathbf{1}_{\mathbf{x}^i_t=\mathbf{m}} \text{log}\left( (\mathbf{x}_0^i)^\top g_{\theta} (\mathbf{x}_0^i |\mathbf{x}^{1:L}_t)\right) \right], \label{eq:elbo}
\end{align}
where $\mathbf{1}_{\mathbf{x}^i_t=\mathbf{m}}$ is the identity function. 
The loss $\Ls(\theta)$ is an upper bound on the negative log likelihood of the data distribution~\citep{shi_simplified_2024, ou_your_2025}. 
$U(0,1)$ is the uniform distribution. 

\subsection{MDLMs in Practice}

\paragraph{Unmasking strategies} 
Equation~\ref{eq:diff_decoding} suggests that the model \emph{randomly} decides---based on the noise schedule $\alpha_t$---whether or not to replace a masked token with the predicted value $f_\theta(\mathbf{x}_t)$.
However, many MDLMs use additional unmasking heuristics that improve the generation quality.  
One approach is to unmask a fixed number of tokens per step, guided not only by the noise schedule but also by the model's confidence.
For example, at time $t$,  Dream-7B~\citep{ye_dream_2025} selects $n\approx L/T$ tokens that have the lowest entropy values. Here, $T$ is the integer number of diffusion steps.
More recent methods introduce exploratory (remasking) and accelerated (aggressive unmasking) decoding stages~\citep{wei_accelerating_2025, wang_remasking_2025}. 
\citet{rutte_generalized_2025} introduce an interpolation between masked and uniform diffusion, which introduces remasking already during the training stage. 

\paragraph{Conditional generation} Conditioning the generative process on a prompt ($\mathbf{c}^{1:L_c}$) is straightforward. 
For decoding, the prompt is simply prefixed to the (partially) masked solution at each iteration, i.e., $\mathbf{p}^{1:L}_{t-1}|\mathbf{c}^{1:L_c} = g_\theta ([\mathbf{c}^{1:L_c}, \mathbf{x}^{1:L}_t])$, where only the last $L$ tokens are updated.

\section{Soft-Masked Diffusion Language Models}\label{sec:method}

As elaborated above, the iterative decoding in MDLMs makes a \emph{binary decision}: selecting either the original mask or the token predicted by the denoising model ($f_\theta$). 
This binary choice results in a loss of valuable contextual information for the masked tokens.
To overcome this limitation, we propose soft-masking (SM). SM augments the mask with intermediate context from the previous denoising step, thereby preserving informative cues and providing a richer input for the next denoising step.

\subsection{Soft-Masking (SM)}
\begin{figure}
    \centering
    \includegraphics[width=.8\linewidth]{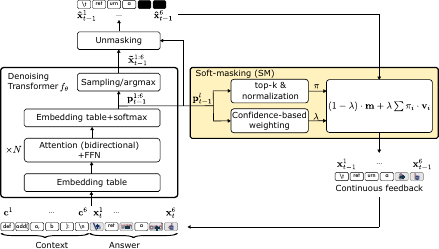}
    \caption{
    Iterative denoising in MDLMs using the proposed soft-masking (SM). 
    Given a context, the aim is to predict the answer via iterative denoising of an initially fully masked response.   
    Here, a bidirectional Transformer ($f_\theta$) performs a single denoising step.  
    This output is passed through an unmasking function that determines which tokens remain masked.
    Our proposed SM enriches the masked tokens by superposing them with the normalized top-$k$ tokens at each position, weighted by a confidence parameter ($\lambda$).
    See Appendix~\ref{app:inference} for an algorithmic description.}
    \label{fig:sm}
\end{figure}

We introduce SM, illustrated in \Figref{fig:sm}, which enhances the denoising process in MDLMs. 
As shown, the overall denoising process follows the standard framework of MDLMs. 
However, instead of discarding previous predictions during masking or remasking, SM gently retains information from the past predictions and incorporates them into subsequent decoding steps.
This provides richer feedback to guide the next denoising step.
To enable this, SM relaxes the binary constraint on the feedback tokens ($\mathbf{x}_{t-1}$) provided to the denoising model, allowing them to represent a \emph{distribution of solutions}, i.e., $\mathbf{x}^l_{t-1} \in \Delta^{|\mathcal{V}|-1}$ instead of the one-hot $\mathbf{x}^l_{t-1} \in \{0,1\}^{|V|}$. 

\paragraph{General formulation}
The feedback is formally defined as:
\begin{align*}
    \mathbf{x}^l_{t-1}=\text{sm}(\mathbf{\hat{x}}^l_{t-1}, \mathbf{p}^l_{t-1}) = \begin{cases} \left(1-\lambda(\mathbf{p}^l_{t-1})\right)\cdot\mathbf{m} + \lambda(\mathbf{p}^l_{t-1}) \sum_{i\in \text{top-$k$}(\mathbf{p}^l_{t-1})} \pi_i \cdot \mathbf{v}_i & \text{if } \mathbf{\hat{x}}^l_{t-1} = \mathbf{m}, \\
\mathbf{\hat{x}}^l_{t-1} & \text{if } \mathbf{\hat{x}}^l_{t-1} \neq \mathbf{m}, 
\end{cases}
\end{align*}
where $\mathbf{p}^l_{t-1}$ is the probability mass vector and $\mathbf{\hat{x}}^l_{t-1}$ is the discrete output from the denoising process. 
SM is applied only to masked tokens; previously predicted tokens remain unchanged. 
SM is implemented as a convex combination of the mask token and a weighted superposition of the top-$k$ predicted tokens. 
Here, $\lambda \in [0,1)$ controls the amount of feedback, $\mathbf{v}_i\in \{0,1\}^{|\mathcal{V}|}$ is a one-hot vector representing token $i$, and $\pi_i$ is the result of the probability mass vector being normalized over the top-$k$ tokens: $\pi_i = [\mathbf{p}^l_{t-1}]_i / \sum_{j \in \text{top-k}(\mathbf{p}^l_{t-1})} [\mathbf{p}^l_{t-1}]_j$, which ensures that $\sum_i \pi_i = 1$.

\paragraph{Confidence-based weighting}
\label{sec:confidence_scaling}
In the following, we describe the dynamic weighting strategy for SM.
Intuitively, a higher confidence should correspond to a greater weight on the model's output, while a lower confidence should preserve more of the original mask token.
To quantify confidence, we use the negative entropy of the probability mass vector $\mathbf{p}^l_{t-1}$, denoted as $-H(\mathbf{p}^l_{t-1})$.
To map this confidence score to a weight in $[0, \omega_s]$, we apply a scaled sigmoid function:
\begin{align}
    \lambda (\mathbf{p}_{t-1}) = \omega_s\cdot \sigma \Big(\omega_a\big(-H(\mathbf{p}^l_{t-1})-\omega_b\big)\Big).
    \label{eq:confidence_scaling}
\end{align}
 where $\sigma(\cdot)$ is the sigmoid function. 
The trainable parameters $\omega_a$, $\omega_b$, and $\omega_s$ control the steepness, the offset, and the amplitude, respectively. 
%
%

\subsection{Learning the SM Feedback}

To teach the richer SM feedback to the model, we introduce a new training methodology that optimizes the SM parameters ($\omega$) concurrently with the main backbone model's parameters ($\theta$). 
This method allows the model to dynamically learn the optimal weighting between the mask token and the predicted tokens for each position, while the backbone simultaneously adapts to the richer feedback. 
As illustrated in Algorithm \ref{alg:training}, the training method is a \emph{two-pass process}. 
Standard MDLMs rely on the analytical tractability of the marginal distribution $q(\mathbf{x}_t|\mathbf{x}_0)$ to enable efficient single-step sampling during training. 
However, the SM introduces a dynamic dependency on the model's intermediate predictions, rendering the exact marginal distribution analytically intractable. 
Our two-pass approach serves as an approximation of this feedback-augmented marginal distribution $\tilde{q}(\mathbf{x}_t|\mathbf{x}_0)$. 
Specifically, we define the effective input state $\mathbf{\tilde{x}}_t = \text{sm}_\omega(\mathbf{x}_t, g_\theta(\mathbf{x}_t))$, where $\mathbf{x}_t \sim q(\cdot|\mathbf{x}_0)$, effectively maximizing the standard variational lower bound (Equation~\ref{eq:elbo}) using this proxy state.

First, we approximate the probability distribution of the previous denoising step by passing the corrupted data through the backbone without a gradient, yielding $\mathbf{\tilde{p}}^{1:L}_{t-1}$. 
This initial pass provides the necessary self-conditioning signal (the soft-masked representation) for the second pass, as practiced in~\citep{chen_analog_2023}.
This distribution is then used to compute the soft-masked representation, which is passed through the backbone for a second time. 
The overall loss $\mathcal{L}$ from this second pass is used to update the learnable parameters for both the backbone ($\theta$) and the SM function ($\omega$) using their respective learning rates, $\eta_{\text{bb}}$ and $\eta_{\text{sm}}$. 
This approach can be highly parallelized with respect to the sequence length, unlike AR training with continuous CoT. 

\citet{arriola_block_2024} showed that a narrower sampling interval for $t$ reduces the gradient variance when optimizing $\Ls$ with batched gradient descent. 
Hence, we sample from the interval $[b_l, b_h]$, $0 \leq b_l < b_h \leq 1$. 
Moreover, following the approach of~\citet{chen_analog_2023}, we activate SM with probability $p_{\text{sm}}\in[0,1]$. 
This prepares the model to cope with both soft-masked and standard inputs, which is particularly necessary at the beginning of the denoising process. 

\begin{algorithm}[t]

    \caption{Training with soft-masking (SM)}
    \label{alg:training}
    \KwIn{Backbone $g_\theta$, training corpus $q_{\text{data}}(\mathbf{x}_0)$, SM function with trainable parameters sm$_\omega$, sampling bounds $0\leq b_l < b_h \leq 1$, learning rates for backbone ($\eta_\text{bb}$) and SM ($\eta_{\text{sm}}$).}
    \KwOut{Trained parameters for backbone ($\theta$) and SM parameters  ($\omega$).}
    \BlankLine
    \Repeat{end training}{
        $\mathbf{x}^{1:L}_0 \sim q_{\text{data}}(\cdot)$ \tcp*{Draw samples from data distribution}
        $t \sim \text{U}(b_l, b_h)$ \tcp*{Draw time step from bounded uniform distribution}
        $\mathbf{x}^{1:L}_t \sim q(\cdot |\mathbf{x}^{1:L}_0)$ \tcp*{Corrupting samples}
        $\tilde{\theta} \leftarrow \text{detach}(\theta)$ \tcp*{Generate detached copy of backbone parameters}
        $\mathbf{\tilde{p}}^{1:L}_{t-1} \leftarrow g_{\tilde{\theta}}\left(\mathbf{x}^{1:L}_t\right)$ \tcp*{First model pass without gradient}
        $\mathbf{p}^{1:L}_{t-1} \leftarrow g_\theta\left(\text{sm}_\omega (\mathbf{x}^{1:L}_{t},\mathbf{\tilde{p}}^{1:L}_{t-1})\right)$ \tcp*{Second model pass with SM and gradient}
        $\mathcal{L}(\theta, \omega) \leftarrow \frac{1}{t} \sum_{l=1}^L \mathbf{1}_{\mathbf{x}^i_t=\mathbf{m}} \text{log}\left( (\mathbf{x}_0^i)^\top \mathbf{p}_{t-1}^i \right) $ \tcp*{Compute loss}
        Update $\theta$ and $\omega$ based on loss $\Ls$ with Adam optimizer using learning rates $\eta_\text{bb}$ and $\eta_\text{sm}$; 
    }
\end{algorithm}

\subsection{SM as an Interpolation Between Absorption and Uniform Diffusion}
This section provides a conceptual interpretation of the proposed SM mechanism. 
To this end, we consider two extreme values of the feedback-scaling parameter ($\lambda=0$ and $\lambda=1$) and simplify the feedback to a single value ($k=1$). 
First, assuming $\lambda=0$ recovers vanilla MDLM. 
The model can always revert to this behavior by setting the scaling factor $\omega_s = 0$.
Second, $\lambda=1$ feeds the previously predicted token (based on argmax) back to the denoising model: 
\begin{align*}
    \text{sm}(\mathbf{\hat{x}}^l_{t-1}, \mathbf{p}^l_{t-1})_{\lambda=1, k=1} =\begin{cases} \mathbf{v}_{\text{argmax}(\mathbf{p_i})} & \text{if } \mathbf{\hat{x}}^l_{t-1} = \mathbf{m}, \\
\mathbf{\hat{x}}^l_{t-1} & \text{if } \mathbf{\hat{x}}^l_{t-1} \neq \mathbf{m}. 
\end{cases}
\end{align*}
A uniform DLM~\citep{austin_structured_2021} would receive the same feedback. However, the unmasking strategy remains active. 
Hence, this corner case can be interpreted as a masked DLM with uniform feedback for the masked states. 
This allows the model to explore different solutions through self-correction, enabled in the masked regions.
Note SM's forward corruption process ($\lambda=1$) deviates from the uniform formulation: SM determines the distribution $q(\mathbf{x}_t|\mathbf{x}_0)$ with the denoising model (see lines 6 and 7 in Algorithm~\ref{alg:training}) rather than from a uniform categorical sampling. 

Relaxing the scaling factor to take intermediate values $\lambda\in[0,1]$ can then be seen as an interpolation between an MDLM and a mask-augmented uniform DLM. 
Importantly, this interpolation occurs in the spatial embedding space.
Why might it be beneficial to retain a portion of the mask token besides attenuating low-confidence predictions?
One reason is that many MDLMs are pretrained to predict masked tokens, and the presence of the mask likely still carries useful positional or structural information. 
This effect is particularly relevant for denoising models that do not use time conditioning.

\begin{table*}[t!]
\caption{
Unconstrained generation after pretraining from scratch.
We report MAUVE ($\uparrow$) and generative perplexity ($\downarrow$) of $L=1024$ generated tokens using MDLM~\citep{sahoo_simple_2024} with binary masking or our SM. Evaluations are tabulated by varying NFE budgets\protect\footnotemark. 
For unmasking, we use either the standard or the more recent ReMDM~\citep{wang_remasking_2025}; the highest scores are bolded. \textit{Gain} shows the performance improvement between the SM and the baseline MDLM. $^\dagger$Results of evaluating the ground-truth data and equal-backbone AR model are taken from~\citep{sahoo_simple_2024}.}
\centering
\label{tab:scratch}
\resizebox{\linewidth}{!}{%
\begin{tabular}{llllrrrrrrrr}
\toprule
 &  & \multirow{2}{*}{Gradient} & \multirow{2}{*}{Forward} & \multicolumn{4}{c}{MAUVE $\uparrow$} & \multicolumn{4}{c}{Generative perplexity $\downarrow$} \\
\cmidrule(r){5-8}\cmidrule(r){9-12}
Unmasking & Feedback & updates  & passes & 1/8 & 1/4 & 1/2 & 1/1 & 1/8 & 1/4 & 1/2 & 1/1 \\
\cmidrule(r){1-4}\cmidrule(r){5-8}\cmidrule(r){9-12}
\multirow{5}{*}{Standard} 
& {Binary} & {1M} & {1M} & {$0.017$} & {$0.025$} & {$0.036$} & {$0.034$}  & {$60.02$} & {$54.95$} & {$52.36$} & {$50.46$}    \\
& {Our SM (iso-compute)} & {0.5M} & {1M} & {$0.143$} & {$\mathbf{0.417}$} & {${0.498}$} & {${0.596}$} & {$41.08$} & {${31.97}$} & {${27.36}$} & {${24.63}$}  \\
& \quad \textit{{Gain}} & & & \gain{+0.126} & \gain{+0.392} & \gain{+0.462} & \gain{+0.562} & \gain{-18.93} & \gain{-22.98} & \gain{-24.99} & \gain{-25.83} \\
& {Our SM (iso-update)} & {1M} & {2M} & {$\mathbf{0.155}$} & {$0.383$} & {$\mathbf{0.535}$} & {$\mathbf{0.602}$} & {$ \mathbf{39.61}$} & {$\mathbf{30.74}$} & {$\mathbf{26.12}$} & {$\mathbf{23.53}$} \\
& \quad {\textit{Gain}} & & & \gain{+0.138} & \gain{+0.358} & \gain{+0.499} & \gain{+0.568} & \gain{-20.41} & \gain{-24.21} & \gain{-26.23} & \gain{-26.93} \\
\cmidrule(r){1-4}\cmidrule(r){5-8}\cmidrule(r){9-12}
\multirow{5}{*}{ReMDM} 
& {Binary} & {1M} & {1M} & {$0.075$} & {$0.199$} & {$0.292$} & {$0.411$} & {$42.53$} & {$31.05$} & {$21.75$} & {$28.62$}     \\
& {Our SM (iso-compute)} & {0.5M} & {1M} & {$\mathbf{0.316}$} & {$\mathbf{0.667}$} & {$\mathbf{0.559}$} & {${0.766}$} & {${29.90}$} & {${18.08}$} & {${11.40}$} & {${17.29}$} \\
& \quad \textit{{Gain}} & & & \gain{+0.241} & \gain{+0.468} & \gain{+0.267} & \gain{+0.355} & \gain{-12.63} & \gain{-12.97} & \gain{-10.35} & \gain{-11.33} \\
& {Our SM (iso-update)} & {1M} & {2M} & {$0.263$} & {$0.626$} & {$0.511$} & {$\mathbf{0.774}$} & {$\mathbf{29.62}$} & {$\mathbf{17.58}$} & {$\mathbf{10.85}$} & {$\mathbf{16.72}$} \\
& \quad {\textit{Gain}} & & & \gain{+0.189} & \gain{+0.427} & \gain{+0.219} & \gain{+0.363} & \gain{-12.91} & \gain{-13.48} & \gain{-10.90} & \gain{-11.90} \\
\cmidrule(r){1-4}\cmidrule(r){5-8}\cmidrule(r){9-12}
\multicolumn{2}{c}{AR ($T=1024$)$^\dagger$} & 0.5M & 0.5M & \multicolumn{4}{c}{0.760} & \multicolumn{4}{c}{12.1}\\ 
\cmidrule(r){1-4}\cmidrule(r){5-8}\cmidrule(r){9-12}
\multicolumn{2}{c}{Data$^\dagger$} & & & \multicolumn{4}{c}{1.0} & \multicolumn{4}{c}{14.8}\\ 
\bottomrule
\end{tabular}
}
\end{table*}

\begin{table*}[h!]
\caption{
MAUVE ($\uparrow$) of unconstrained generation after pretraining continuation. 
\textit{Gain} shows the performance improvement between the SM and the binary MDLM with pretraining continuation.}
\centering
\label{tab:mauve}
\resizebox{\linewidth}{!}{%
\begin{tabular}{lllllll}
\toprule
 &  & \multirow{2}{*}{Gradient} & \multicolumn{4}{c}{NFE budget} \\
\cmidrule(r){4-7}
Unmasking & Feedback & updates  & 1/8 & 1/4 & 1/2 & 1/1 \\
\cmidrule(r){1-3}\cmidrule(r){4-7}
\multirow{5}{*}{Standard} 
& Binary & 1M & {$0.017$} & {$0.025$} & {$0.036$} & {$0.034$}      \\
& Binary & 1M+100k & {$0.018_{(\pm0.000 )}$} & {$0.027_{(\pm0.005 )}$} & {$0.032_{(\pm0.003 )}$} & {$0.038_{(\pm0.002 )}$}  \\
& {Our SM (iso-compute)} & {1M+50k} & {$0.054_{(\pm0.009 )}$} & {$0.129_{(\pm0.029 )}$} & {$0.200_{(\pm0.038 )}$} & {$\mathbf{0.259}_{(\pm0.024 )}$}  \\
& \quad {\textit{Gain}} & & \gain{+0.036} & \gain{+0.101} & \gain{+0.168} & \gain{+0.221} \\
& Our SM (iso-update) & 1M+100k & {$\mathbf{0.059}_{(\pm 0.007 )}$} & {$\mathbf{0.139}_{(\pm 0.021)}$} & {$\mathbf{0.232}_{(\pm 0.026 )}$} & {$0.211_{(\pm 0.145 )}$}
   \\
& \quad \textit{Gain} & & \gain{+0.041} & \gain{+0.112} & \gain{+0.200} & \gain{+0.173} \\
\cmidrule(r){1-3}\cmidrule(r){4-7}
\multirow{5}{*}{ReMDM} 
& Binary & 1M & {$0.075$} & {$0.199$} & {$0.292$} & {$0.411$}    \\
& Binary & 1M+100k & {$0.052_{(\pm 0.005 )}$} & {$0.180_{(\pm 0.030 )}$} & {$0.315_{(\pm 0.032 )}$} & {$0.421_{(\pm 0.021 )}$}  \\
& {Our SM (iso-compute)} & {1M+50k} & {$0.137_{(\pm0.011 )}$} & {$\mathbf{0.441}_{(\pm 0.064)}$} & {$0.610_{(\pm0.020 )}$} & {$\mathbf{0.693}_{(\pm0.033)}$}  \\
& \quad \textit{Gain} & & \gain{+0.084} & \gain{+0.262} & \gain{+0.295} & \gain{+0.272} \\
& Our SM (iso-update) & 1M+100k  & {$\mathbf{0.146}_{(\pm 0.014 )}$} & {$0.432_{(\pm 0.035 )}$} & {$\mathbf{0.617}_{(\pm 0.020 )}$} & {$0.692_{(\pm 0.034 )}$}  \\
& \quad \textit{Gain} & & \gain{+0.094} & \gain{+0.252} & \gain{+0.302} & \gain{+0.271} \\
\bottomrule
\end{tabular}
}
\end{table*}

\section{Experiments}\label{sec:experiments}
\paragraph{General setup} 
We begin by evaluating SM on a small-scale language modeling benchmark, using a 169M-parameter MDLM~\citep{sahoo_simple_2024} to demonstrate its benefits with both standard and improved unmasking strategies.
We then apply SM to the large-scale Dream-7B~\citep{ye_dream_2025} and Dream-Coder-7B~\citep{xie_dreamcoder_2025} models, showing improvements on downstream coding tasks.
In addition to training from scratch, we assess the efficiency of adapting existing models via continued pretraining (for small-scale models) or finetuning (for large-scale models). 
In the pretraining continuation and finetuning setup, the baseline models with binary masking are trained with the same procedure for a fair comparison. 
Crucially, since our proposed training algorithm (Alg.~\ref{alg:training}) requires two model forward passes per iteration (versus one in the standard training), we evaluate SM under two distinct computational budgets:
(1) \textbf{Iso-update:} We match the total number of gradient updates ($N$). This isolates learning efficiency but requires roughly twice the wall-clock time for SM.
(2) \textbf{Iso-compute:} We match the total number of model forward passes. In this setting, SM is trained for $N/2$ number steps, ensuring the total computational cost remains equivalent to the baseline.
%
\footnotetext{The NFE budget (Appendix~\ref{app:nfe_budget}) is the ratio between the diffusion steps and the max generation length.}
\subsection{Language Modeling}
\label{sec:language_modeling}
\paragraph{Setup} 
We evaluate SM on language modeling using a 169M-parameter MDLM~\citep{sahoo_simple_2024} trained on the OpenWebText (OWT)~\citep{Gokaslan2019OpenWeb}. 
The model utilizes a Diffusion Transformer (DiT) backbone~\citep{peebles_scalable_2023}, which integrates time-step conditioning into an encoder-only transformer.
We investigate two training regimes for SM (with $k=3)$: (1) pretraining from scratch for up to 1M steps, and (2) efficient adaptation, where we apply continued pretraining to a pretrained binary MDLM for an additional 100k steps.
For evaluation, we report perplexity on the OWT validation set (computed via the two-pass SM objective). 
As a second measure, we assess the unconstrained generation quality using both generative perplexity and the MAUVE score~\citep{pillutla2021mauve}, the latter serving as a robust metric for diversity and quality.
%
%
%
We vary the number of function evaluations (NFE) between 128 and 1024 (representing $1/8$ to $1/1$ of the compute budget). 
Both the baseline MDLM and our MDLM with SM require the same number of model passes. 
In addition to MDLM's standard unmasking, we also evaluate the models with a recent remasking strategy (ReMDM;~\citealt{wang_remasking_2025}). 
See Appendix~\ref{app:finetuning-mdlm} for more details.

\begin{figure}[t!]
    \centering 
    \begin{subfigure}{0.45\textwidth}
        \centering
        \includegraphics[width=\textwidth]{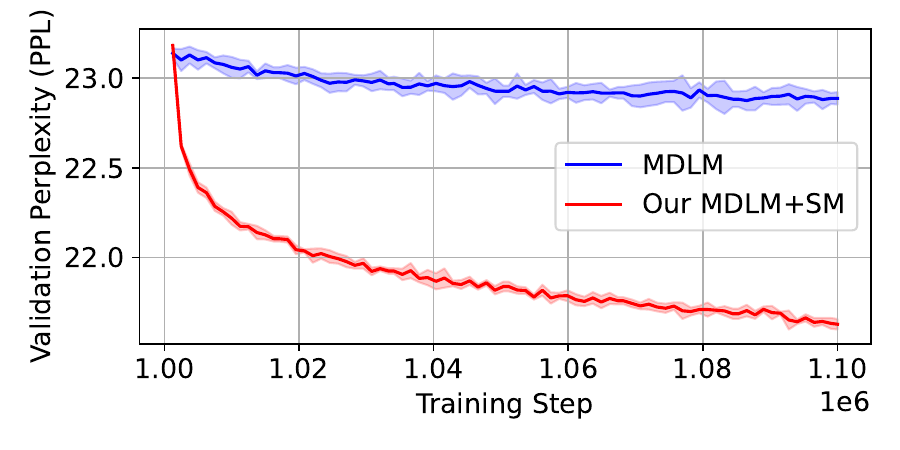}
        \caption{Validation perplexity on OWT.}
        \label{fig:owt-perplexity}
    \end{subfigure}
    \hfill 
    \begin{subfigure}{0.45\textwidth}
        \centering
        \includegraphics[width=\textwidth]{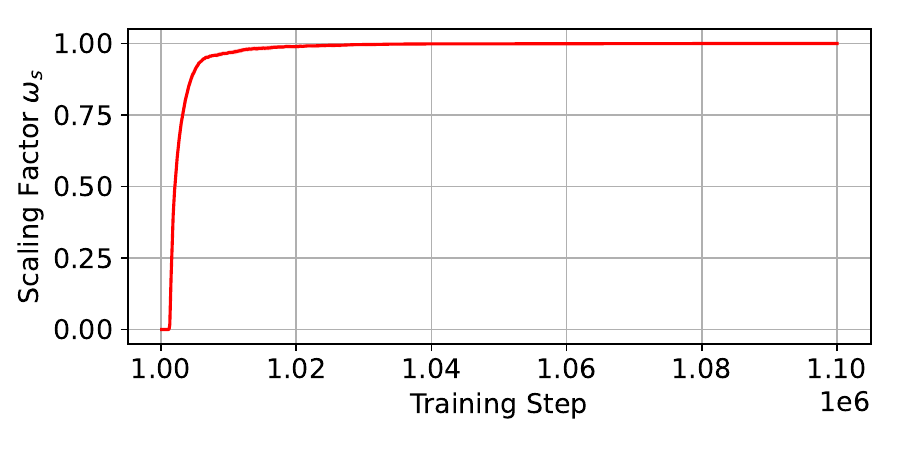}
        \caption{SM's learned scaling factor ($\omega_s$).}
        \label{fig:owt-scale}
    \end{subfigure}
    \caption{Continuing MDLM pretraining on OpenWebText. We show the average $\pm$ standard deviation (shaded) across 5 seeds. SM is configured with $k=3$. (a) Our SM yields better (lower) validation perplexity than binary masking. (b) The model learns to fully use SM by increasing its influence over the scaling factor $\omega_s$.}
    \label{fig:owt}
\end{figure}

\paragraph{Results}
Table~\ref{tab:scratch} shows that our SM model trained from scratch, both under the iso-compute and iso-update budgets, consistently improves the MAUVE score (by up to +0.568 points) and the generative perplexity (by up to -26.93 points) when using standard unmasking. 
Interestingly, the iso-compute SM model (with $N/2$=0.5M pretraining steps) even slightly outperforms the iso-update model at certain lower NFE budgets (e.g., NFE 1/4). 
Thus, we observe that SM is particularly effective in compute-restricted training regimes.
Moreover, SM can further benefit from advanced remasking strategies, surpassing ReMDM with the binary MDLM (by up to +0.468 MAUVE points), and even outperforming the AR MAUVE score (0.760) at the highest compute budget (achieving 0.774).
Table~\ref{tab:scratch-entropy} in Appendix~\ref{app:entropy} also shows that SM maintains the entropy. 
Additionally, our MDLM with SM achieves superior OWT validation perplexities (21.47 in iso-update and 22.36 in iso-compute), as shown in Appendix~\ref{app:val-perplexity}.

Next, we demonstrate that SM can be efficiently integrated into a pre-existing binary MDLM via continued pretraining for up to 100k steps. 
As shown in~\Figref{fig:owt-perplexity}, our SM decreases the validation perplexity on OWT (from 23.14 to 21.63).
Binary MDLM (without SM) also improves the perplexity, but by a much smaller margin (from 23.14 to 22.88).
\Figref{fig:owt-scale} shows that MDLM learns to make use of the richer SM feedback by increasing the scale from initially near-zero to close to 1. 
The gain in validation perplexity transfers to unconstrained generation (Table~\ref{tab:mauve}) in pretraining continuation too, where SM consistently improves the MAUVE score across all NFE budgets and unmasking strategies.
Finally, all observations in MAUVE score transfer to generative perplexities and entropy, as shown in Appendix~\ref{app:entropy}.

\paragraph{Ablations}
We evaluate our SM design choices on language modeling in Appendix~\ref{app:ablation-lm}. 
We find that using SM 80\% of the time and $k=3$ superposition yields the best validation perplexities.  
Moreover, an alternative SM feedback with softmax (with a trainable temperature) instead of top-k achieves competitive validation perplexities, at the cost of higher compute and memory demands. 
Even though the softmax allows for propagating the gradients to the first model pass, applying the update based on both model passes does not improve the perplexities while increasing the computational cost. 
Finally, Appendix~\ref{app:speed} shows that the overhead in inference by SM is small (12\%).

\begin{table}[t!]
\caption{
Accuracy (\%) on coding tasks. 
Evaluations are tabulated by varying NFE budgets.
We finetune the models with 5 seeds and report the mean accuracy ($\pm$ standard deviation). SM is configured with $k$=1.
\textit{Gain} shows the comparison between the SM model and the finetuned baseline. The best performing model is marked in bold. The learned SM parameters are given in Appendix~\ref{app:learned_parameters}.
}
\label{tab:downstream}
\resizebox{\linewidth}{!}{%
\begin{tabular}{clcllllllll}
\toprule
\multicolumn{1}{l}{}                                 &          & \multicolumn{1}{l}{}                                       & \multicolumn{4}{c}{Dream-Coder-7B (instruct)} & \multicolumn{4}{c}{Dream-7B (instruct)} \\
\cmidrule(r){4-7}\cmidrule(r){8-11}
\begin{tabular}[c]{@{}c@{}}NFE\\ budget\end{tabular} & Feedback & \begin{tabular}[c]{@{}c@{}}FT\\ steps\end{tabular} & HumanEval    & HumanEval+    & MBPP    & MBPP+   & HumanEval     & HumanEval+     & MBPP    & MBPP+    \\
\cmidrule(r){1-3}\cmidrule(r){8-11}\cmidrule(r){4-7}\cmidrule(r){8-11}
\multirow{4}{*}{1/4}                                 & Binary   & -                                                          &  $25.0$&        $25.0$&         $27.4$&         {$29.4$}&        $18.9$&         $17.1$&         $26.6$&          {$30.2$}\\
                                                     & Binary   & 33.5k                                                      &       $28.5_{(\pm 1.3)}$&        $27.7_{(\pm 1.8)}$&         $25.9_{(\pm 1.5)}$&         {$24.6_{(\pm 1.7)}$}&        $19.0_{(\pm1.7)}$&         $15.9_{(\pm 2.8)}$&         $27.0_{(\pm 1.6)}$&          {$29.2_{(\pm 1.5)}$}\\
                                                     & Our SM       & 33.5k                                                      &       $\mathbf{29.5}_{(\pm 1.8)}$&        $\mathbf{28.2}_{(\pm 1.7)}$&         $\mathbf{33.2}_{(\pm 1.8)}$&         {$\mathbf{29.4}_{(\pm 1.9)}$}&        $\mathbf{24.8}_{(\pm1.8)}$&         $\mathbf{23.0}_{(\pm1.3)}$&         $\mathbf{32.3}_{(\pm1.3)}$&          {$\mathbf{36.7}_{(\pm1.0)}$}\\
                                                     & \quad \textit{Gain}     &                                                            &   \gain{+1.0}&        \gain{+0.5}&         \gain{+7.3}&         \gain{+4.8}&        \gain{+5.8}&         \gain{+7.1}&         \gain{+5.3}&    \gain{+7.5}\\
                                                     \cmidrule(r){1-3}\cmidrule(r){8-11}\cmidrule(r){4-7}\cmidrule(r){8-11}
\multirow{4}{*}{1/2}                                 & Binary   & -                                                          &  $54.9$&        $50.6$&         $51.6$&         {$51.3$}&        $31.1$&         $29.3$&         $42.8$&          {$45.8$}\\
                                                     & Binary   & 33.5k                                                      &       $53.8_{(\pm 1.4)}$&        $49.3_{(\pm 1.6)}$&         $49.8_{(\pm 0.9)}$&         {$53.2_{(\pm 1.5)}$}&        $33.0_{(\pm3.0)}$&         $29.5_{(\pm3.4)}$&         $43.1_{(\pm0.4)}$&          {$39.6_{(\pm2.7)}$}\\
                                                     & Our SM       & 33.5k                                                      &       $\mathbf{57.2}_{(\pm 2.7)}$&        $\mathbf{52.6}_{(\pm 2.0)}$&         $\mathbf{56.2}_{(\pm 0.7)}$&         {$\mathbf{56.4}_{(\pm 1.4)}$} &        $\mathbf{38.3}_{(\pm1.9)}$&         $\mathbf{33.8}_{(\pm2.6)}$&         $\mathbf{48.4}_{(\pm1.2)}$&          {$\mathbf{54.7}_{(\pm1.8)}$}\\
                                                     & \quad \textit{Gain}     &                                                            &   \gain{+3.4}&        \gain{+3.3}&         \gain{+6.4}&         \gain{+3.2}&        \gain{+5.3}&         \gain{+4.3}&         \gain{+5.3}&    \gain{+15.1}\\
                                                     \cmidrule(r){1-3}\cmidrule(r){8-11}\cmidrule(r){4-7}\cmidrule(r){8-11}
\multirow{4}{*}{1/1}                                 & Binary   & -                                                          &  $75.0$&        $69.5$&         $65.8$&         {$70.4$}&        $57.9$&         $53.0$&         $57.8$&          {$63.5$}\\
                                                     & Binary   & 33.5k                                                      &       $75.7_{(\pm 1.7)}$&        $68.9_{(\pm 2.0)}$&         $65.6_{(\pm 0.8)}$&         {$68.1_{(\pm 1.1)}$}&        $\mathbf{59.5}_{(\pm1.8)}$&         $\mathbf{53.0}_{(\pm1.0)}$&         $\mathbf{58.3}_{(\pm0.1)}$&          {$\mathbf{62.8}_{(\pm0.7)}$}\\
                                                     & Our SM       & 33.5k                                                      &       $\mathbf{76.2}_{(\pm 1.4)}$&        $\mathbf{70.4}_{(\pm 1.3)}$&         $\mathbf{67.0}_{(\pm 0.7)}$&         {$\mathbf{69.6}_{(\pm 0.9)}$}&        $57.8_{(\pm1.9)}$&         $50.0_{(\pm0.7)}$&         $56.4_{(\pm1.2)}$&          {${61.9}_{(\pm0.8)}$}\\
                                                     & \quad \textit{Gain}     &                                                            &   \gain{+0.5}&        \gain{+1.5}&         \gain{+1.4}&         \gain{+1.5}&        \textit{-1.7}&         \textit{-3.0}&         \textit{-1.9}&    \textit{-0.9}\\
\bottomrule
\end{tabular}
}
\end{table}

\subsection{Code Generation}
\label{sec:code_generation}
\paragraph{Setup} 
We integrate SM into the state-of-the-art Dream-7B~\citep{ye_dream_2025} and Dream-Coder-7B~\citep{xie_dreamcoder_2025} instruction-tuned models. 
For finetuning, we aim to use the same SFT datasets as the original models. 
Dream-7B uses Tulu 3~\citep{tulu3_2025} and SmolLM2~\citep{smol_2025}; Dream-Coder-7B uses Ling-Coder-SFT~\citep{codefuse2025samplemattersleveragingmixtureofexperts}.
We deploy parameter-efficient finetuning using weight-decomposed low-rank adaptation (DoRA;~\citealt{shih_dora_2024}) on nearly 270k curated training samples with a batchsize of 8, yielding 33.5k update steps. 
We test the models on two coding tasks, HumanEval~\citep{chen_evaluating_2021} and MBPP~\citep{mbpp_2021}, as well as on their plus version~\citep{evalplus_2023}. 
{
We report results in iso-update training, and show similar gains in iso-compute in Appendix~\ref{app:coding-isocompute}. 
}
See Appendix~\ref{app:finetuning-large} for more details on the experimental setup.

\begin{wrapfigure}[13]{R}{0.66\textwidth}
    \vspace{-18pt}
    \centering
    \centering 
    \begin{subfigure}{0.32\textwidth}
        \centering
        \includegraphics[width=\textwidth]{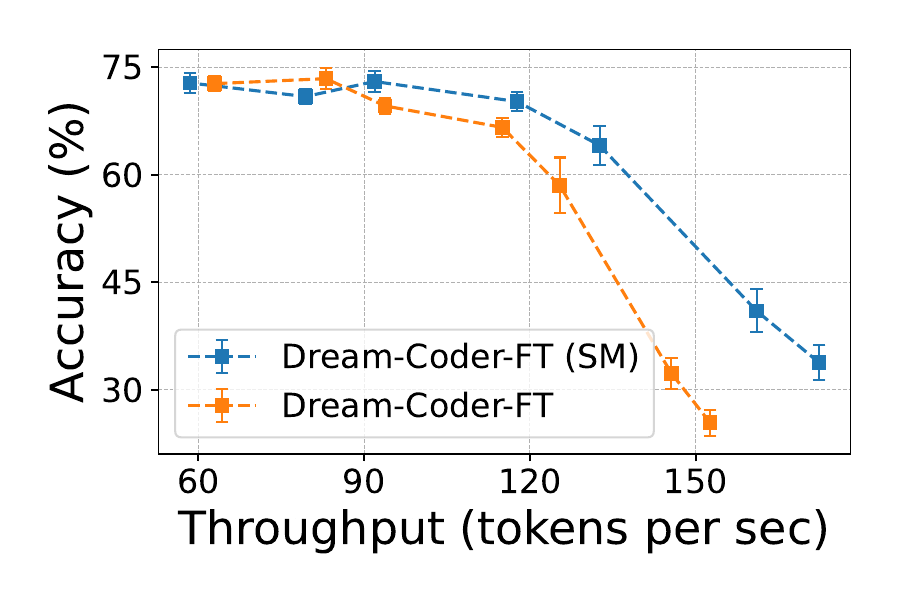}
        \caption{HumanEval}
        \label{fig:fast_dllm_he}
    \end{subfigure}
    \hfill 
    \begin{subfigure}{0.32\textwidth}
        \centering
        \includegraphics[width=\textwidth]{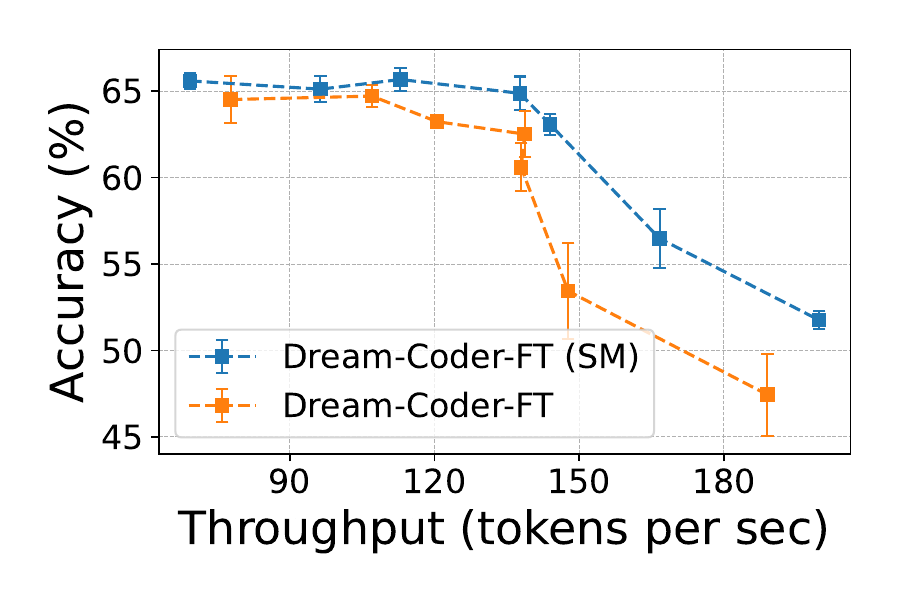}
        \caption{MBPP}
        \label{fig:fast_dllm_mbpp}
    \end{subfigure}
    \caption{Integrating SM into Fast-dLLM. We plot Dream-Coder-7B performance vs. throughput with both binary feedback and our SM. SM again excels in high-throughput settings.}
    \label{fig:fast_dllm}
\end{wrapfigure}
\paragraph{Results}
The mean results are given in Table~\ref{tab:downstream}, tabulated based on the NFE budget of the decoding. 
First, we see that finetuned models (without SM) can maintain the performance of the original models, validating our finetuning procedure. 
Second, the benefits observed in small-scale language modeling transfer to performance gains in large-scale models: the table shows a performance boost on nearly all tasks (up to 15.1\%). The gains are particularly prominent at lower NFE budgets.
Furthermore, we show that SM complements other efficiency-enhancement mechanisms. Figure~\ref{fig:fast_dllm} reports the mean performance of the finetuned Dream-Coder-7B models, with and without SM, when combined with Fast-dLLM's blockwise caching and confidence-aware decoding~\citep{wu_fastdllm_2026}. The performance is plotted as a function of the token throughput, which is indirectly determined from the block length of the blockwise decoding---with longer blocks correlating with higher throughputs, as described in Appendix~\ref{app:fast_dllm}. The benefits of SM become more promising at higher throughputs.

\paragraph{Ablations}
To validate our design choices, we perform extensive ablation studies on the Dream models. 
In Appendix~\ref{app:math_reasoning}, we evaluate SM on math tasks and notice similar performance gains.
In Appendix~\ref{app:top-k}, we vary the number of top-$k$ contributions. 
We find that $k=1$ is optimal for coding tasks and $k=3$ yields the best overall results.  
We also tested a trainable k selection, by replacing the top-$k$ filtering with a softmax that uses a trainable temperature. 
However, this did not improve the performance. 
Finally, Appendix~\ref{app:time} applies SM only in certain periods during denoising. 
We find that SM is particularly beneficial in the first 20\% of decoding steps.

\section{Related Works}

\paragraph{Continuous feedback in AR}
COCONUT~\citep{hao_training_2025} feeds continuous token predictions back into the model to enhance reasoning capabilities.
This requires training from scratch---an inherently sequential process due to its reliance on previous continuous token outputs.
To reduce computational demands, many have proposed summarizing sequences of tokens into higher-level continuous representations or concepts~\citep{lcm_large_2024, tack_llm_2026, geiping_scaling_2025}.
Approximate training methods such as Jacobi iterations~\citep{wu_parallel_2025} further mitigate sequential training bottlenecks by iteratively refining AR thought tokens.
More recent methods feed weighted superpositions of token predictions back into the model without additional training~\citep{zhuang_mixture_2025, zhang_soft_02}, enabling lightweight and training-free continuous feedback.
While these approaches can improve task performance, AR models often suffer from unreliable halting behavior, necessitating entropy-based heuristics to terminate decoding.
In contrast, our SM introduces continuous feedback directly into MDLMs, preserving parallelism during training and inference, incurring only constant overhead during training, and avoiding halting heuristics.

\paragraph{Discrete vs. continuous representations in diffusion modeling}
Continuous DLMs maintain a continuous latent space throughout the diffusion process and perform a discretization step at the final readout~\citep{li_diffusionlm_2022, dieleman_continuous_2022, gong_diffuseq_2022,strudel_selfconditioned_2022, gong_diffuseqv2_2023, gulrajani_likelihoodbased_2023}; however, they generally achieve lower performance and do not allow for adaptation from pretrained AR models. 
Several works explore hybrid representations.
HART~\citep{tang_hart_2025} augments a discrete AR-based image predictor with a residual continuous diffusion model to correct quantization errors, but remains dependent on unidirectional AR generation, limiting self-correction.
Self-conditioning~\citep{chen_analog_2023} is a closely related approach that uses a similar two-pass training methodology in training. 
However, their use of concatenation increases the model complexity due to the resultant higher input dimensionality, a critical architectural difference from our method. 
Furthermore, it does not inherently offer a mechanism for a smooth adaptation.
%
\citet{sahoo_diffusion_2025} derive discrete uniform-state diffusion from continuous Gaussian models, enabling faster training and generation, though scalability and downstream performance remain unproven.
%
\citet{chao_masked_2025} propose fine-grained token representations using $l$-dimensional vectors with base-$b$ values, allowing partial unmasking during denoising, which is most effective with many decoding steps ($T \gg L$).
In contrast, our SM approach improves decoding performance while maintaining constant input dimensionality. By superposing the \mask token and top-$k$ predictions, SM introduces continuous feedback without increasing complexity.

\paragraph{Efficiency improvements of MDLMs}
%
dLLM-Cache~\citep{liu_dllmcache_2025} introduces caching to MDLMs, maintaining an almost static cache for prompts and a dynamic cache for the responses. 
This yields a speedup of up to 9$\times$ for long prompts at iso-accuracy. 
Semi-autoregressive generation via block diffusion~\citep{arriola_block_2024, nie_large_2025}, optionally combined with caching~\citep{wu_fastdllm_2026}, offers speedups but compromises full bidirectionality.
NFE efficiency and generation quality have also been improved through dynamic unmasking strategies~\citep{jin_thinking_2025, wei_accelerating_2025, wang_remasking_2025}, which adapt the masking schedule during decoding.
These techniques are complementary to our SM approach and can be readily integrated, as we have already demonstrated by integrating ReMDM~\citep{wang_remasking_2025} and Fast-dLLM~\citep{wu_fastdllm_2026} with SM. 

\paragraph{Concurrent works}
Concurrently with the development of SM, several other works~\citep{zheng_continuously_2026, pynadath_candi_2025, jo_loopholing_2026, zhou_coevolutionary_2025, shariatian_latent_2025, ma_dinfer_2025} sought to address information loss in MDLMs. Appendix~\ref{app:concurrent} provides an elaborate discussion and comparison, demonstrating that SM achieves superior language modeling performance where such comparisons are feasible.

\section{Conclusion}
We introduced soft-masking (SM), a lightweight mechanism for incorporating continuous feedback into masked diffusion language models (MDLMs).
By blending the \mask token with a convex combination of top-$k$ predictions during iterative decoding, SM enables more expressive and flexible updates without increasing model complexity.
Applied to both small and large-scale MDLMs, SM consistently improves performance across language modeling and coding tasks. 
These results demonstrate that continuous feedback can enhance the capabilities of discrete diffusion models.

\paragraph{Limitations and future works}

Even though the training with SM is parallelizable in the sequence length, it requires an additional forward pass, which increases the complexity. 
We see future work in incorporating reinforcement learning-based methods~\citep{black_training_2023, zhao_d1_2025, zhu_llada_2025} to leverage the richer feedback. 
\section*{Ethics Statement}
This work does not involve human subjects, personally identifiable information, or sensitive data. All experiments were conducted using publicly available models and datasets. The proposed method is intended for research in code generation and was evaluated in controlled settings. Our focus on improving model performance in low-compute budget environments was a central consideration to mitigate environmental impact. The authors declare no known conflicts of interest.

\section*{Reproducibility Statement}
This paper describes the proposed SM method, including the training Algorithm~\ref{alg:training} and inference Algorithm~\ref{alg:inference}. 
The setup used for training and evaluating our model on language modeling tasks is described in Appendix~\ref{app:finetuning-mdlm}. 
The setup used for training and evaluating the Dream models for code generation is available in Appendix~\ref{app:finetuning-large}. 
The code for both language modeling and code generation is provided in the supplementary materials, along with the required experimental environments.

\section*{Acknowledgments}
This work was partially supported by the European Union's Horizon Europe Research and Innovation Program under Grant Agreement No. 101223271, and by the Swiss State Secretariat for Education, Research and Innovation (SERI) under contract number 25.00443.
We thank Ronan Tanios for his contributions to the experimental evaluation.
Moreover, we are grateful to Zaira Nazario and Abu Sebastian for the managerial support.



\bibliography{references_hd_learning, bib_short}
\bibliographystyle{iclr2026_conference}

\newpage
\tableofcontents 
\newpage
\appendix
\section{Inference with SM}\label{app:inference}
Algorithm~\ref{alg:inference} describes the inference procedure with SM. 

\begin{algorithm}[t]
    \caption{Inference with soft-masking (SM)}
    \label{alg:inference}
    \KwIn{Backbone $g_\theta$, SM function with parameters sm$_\omega$, generation length $L$, number of denoising steps $T$. }
    \KwOut{Generated sequence $\mathbf{\hat{x}}_0^{1:L}$.}
    \BlankLine
    $\mathbf{x}^{1:L}_T \leftarrow \mathbf{m}, \mathbf{m}, ..., \mathbf{m} $ \tcp*{Initialize sequence with full masks}
    \Repeat{t=0}{
        $\mathbf{p}^{1:L}_{t-1} \leftarrow g_\theta\left(\mathbf{x}^{1:L}_{t}\right)$ \tcp*{Backbone pass with SM}
        $\mathbf{\tilde{x}}^{1:L}_{t-1} \leftarrow \text{sample}(\mathbf{p}^{1:L}_{t-1}) $\tcp*{Sample from backbone distribution (e.g., nucleus)}
        $\mathbf{\hat{x}}^{1:L}_{t-1} \leftarrow \text{unmask}(\mathbf{\tilde{x}}^{1:L}_{t-1}, \mathbf{p}^{1:L}_{t-1}, t, T, L) $\tcp*{Unmasking (e.g., based on entropy)}
        $\mathbf{x}^{1:L}_{t-1} \leftarrow \text{sm}_\omega (\mathbf{\hat{x}}^{1:L}_{t},\mathbf{{p}}^{1:L}_{t})$ \tcp*{Computing SM feedback}
         $t \leftarrow t-1$ 
    }
\end{algorithm}
\section{Experimental setup}
This appendix provides details on the experiments conducted for language modeling and code generation. 
All experiments were run on 1--2 compute nodes with 1--8 NVIDIA A100 GPUs, each with 80~Gigabytes of VRAM.

\subsection{Language Modeling}\label{app:finetuning-mdlm}

\paragraph{Pretraining from scratch}
The pretraining experiments follow the setup by~\cite{sahoo_simple_2024}. 
We use a bidirectional Transformer backbone with 12 layers, 12 attention heads, and 768 hidden dimensions. 
The model is tokenized using the GPT-2 tokenizer~\citep{radford_2019_gpt2}.
Pretraining is performed on the same OpenWebText (OWT)~\citep{Gokaslan2019OpenWeb} split, with the last 100k documents reserved for validation.
We use an AdamW optimizer with a linear learning warm-up for the first 2500 steps, and then keeping it constant at $\eta_{\text{bb}}=$3e-4 and $\eta_{\text{sm}}=$1e-2 for the backbone and the SM parameters, respectively. 
We train the model for 1M training steps using a batchsize of 512, which yields 262B tokens seen during training. 
The dropout rate is set at 0.1.
Training was performed on 2 compute nodes, each with 8 A100 GPUs (80~Gigabytes of VRAM each), using a batchsize of 32 per device and deploying gradient accumulation to achieve an effective batchsize of 512. 
%

\paragraph{Pretraining continuation}
Our starting checkpoint was pretrained on OWT for 1M steps and was released by~\cite{sahoo_simple_2024}. 
We train the model for 100k training steps using the same hyperparameters as in pretraining from scratch.
We train each model with 5 different seeds (1, 2, 3, 4, 5) to account for training variability. 
Training was performed on 4 A100 GPUs (80~Gigabytes of VRAM each) using a batchsize of 32 per device and deploying gradient accumulation to achieve an effective batchsize of 512. 
Pretraining one model took approximately 64 hours and 139 hours for binary and SM, respectively.

\paragraph{Soft-masking parameterization}
\label{app:parameterization}For the SM feedback, we add a trainable module to the model. 
This module contains all SM logic and augments the input embeddings with SM before the main forward process. 
We initialize the three SM-feedback parameters with the parameters given in Appendix~\ref{app:initialization}.
We have a few imposed constraints on these values: $\omega_s \in [0,1], \omega_a \geq 0$, and $\omega_b \leq 0$.
To account for these constraints, we apply simple re-parameterizations during training:
\begin{itemize}
    \item $\omega_s$ is passed through a sigmoid, ensuring it remains in $[0,1]$.
    \item $\omega_a$ and $\omega_b$ (negative version) are each passed through a softplus, guaranteeing non-negativity.
\end{itemize}
Before inference, the learned parameters are de-parameterized: we take the direct output of the optimization, apply the inverse transforms, and for $\omega_b$ additionally negate the result so that it respects the $\omega_b \leq 0$ constraint.
All other SM parameters (i.e., $k$) are specified in our added module. 
We simply perform a forward pass through this module to get the mixing weights for the new input embeddings.  
    
\paragraph{Generative perplexity, entropy, and MAUVE in unconstrained generation}
Our unconstrained generation evaluation follows the experimental setup by~\cite{wang_remasking_2025}. 
We perform unconstrained generation of 5000 samples ($L=1024)$ per model with a batchsize of 1 using nucleus sampling with $p=0.9$. 
We use GPT-2 large for measuring the generative perplexity. 
Moreover, we use GPT-2 large as the embedding model for MAUVE score computation, where we set the MAUVE scaling hyperparameter to 5. 
Concerning ReMDM unmasking, we use the max-capped schedule ($\eta_{\text{cap}}=0.04$) for fast sampling ($T<L$) and the loop-strategy ($t_{\text{on}}=0.55, t_{\text{off}}=0.05,\alpha(t_{\text{on}})=0.9, \eta_{\text{cap}}=0.02$) for inference-time scaling ($T\geq L$). 

\subsection{Code Generation}\label{app:finetuning-large}
This appendix describes the details of our finetuning experiments on Dream-7B and Dream-Coder-7B. 

\paragraph{Backbone models} 
\label{sec:backbones}
We use the pretrained Dream-7B~\citep{ye_dream_2025} and Dream-Coder-7B~\citep{xie_dreamcoder_2025} backbones, respectively. 
Both of these are 7B parameter models that are adapted from the Qwen2.5 family~\citep{qwen_qwen25_2025, hui_qwen25coder_2024}.
For both models, we use the instruction-tuned versions.

\paragraph{Tasks} We primarily evaluate on four code synthesis benchmarks:
\begin{itemize}
    \item \textbf{HumanEval}~\cite{chen_evaluating_2021}: a benchmark of 164 Python programming problems designed to test a model’s ability to write correct and functional code from natural language specifications.
    \item \textbf{MBPP}~\cite{mbpp_2021}: a benchmark consisting of 974 “Mostly Basic Programming Problems,” each specified in natural language and accompanied by input–output test cases, targeting introductory-level programming tasks. We only perform evaluation with the 500 samples in the test subset. It is important to note that some works report higher scores for the Dream models on the MBPP task. This is a result of using the ``sanitized" subset of the data, instead of the standard lm-eval version that we used.
    \item \textbf{EvalPlus: HumanEval$\mathbf{+}$ and MBPP$\mathbf{+}$}~\cite{evalplus_2023}: Extended benchmarks involving adding more unique test cases and correcting any inaccurate ground-truth solutions. 
\end{itemize}

For our experiments, we use the standard \path{instruct} implementations of these benchmarks as implemented by \path{lm-evaluation-harness} at the time of writing.  
For HumanEval$+$, we created a custom \path{instruct} version of the task, using the same prompt as \path{humaneval_instruct}.
All tasks were evaluated in a zero-shot setting with a temperature of $0.1$, a top-p value of $0.9$, and the entropy-based unmasking algorithm.
The HumanEval tasks were evaluated with a max generation length of 768, and the MBPP tasks were evaluated with a generation length of 512.
\paragraph{NFE Budget}
\label{app:nfe_budget}

MDLMs typically have a maximum generation length as a parameter during generation. For our experiments, we use 768 for HumanEval(+) and 512 for MBPP(+).
These models also have a parameter that quantifies the number of diffusion steps that should be taken to unveil all tokens.
Typically, diffusion models set the same number of steps as the number of maximum tokens. 
In these cases, exactly one token is unmasked at each diffusion step. 

Decreasing the number of diffusion steps leads to a much more efficient computation---by unmasking more than one token each step. 
From a computational cost perspective, this is essentially a linear relationship. 
If four tokens are unmasked at each denoising step, the generation will happen $4\times$ faster than if only one is sampled each step.
This leads us to define the NFE budget of a generation:

Given a fixed-length generation task with a max number of tokens, the NFE budget of the generation will be:
\begin{align}
    \text{NFE budget} = \frac{\text{\# of diffusion steps}}{\text{max \# of tokens}}
    \label{eq:comp_budget}
\end{align}

In our experiments, we discuss NFE budgets of 1/4, 1/2, and 1/1. If the NFE budget is 1/$n$, it means that, on average, $n$ tokens are unmasked per step. 

\paragraph{General setup and hyperparameters}
We aimed to keep the finetuning implementation as close to the original Dream-7B~\citep{ye_dream_2025} SFT implementation as possible. 
Due to the fact that the exact SFT implementation code is unavailable, we use the DiffuLLaMa code~\citep{gong_scaling_2025} and the Dream paper~\citep{ye_dream_2025} for reference. 
Algorithm~\ref{alg:training} displays an overview of the algorithm that we use. 
The only change for scaling the algorithm for finetuning larger models is that, rather than updating the full weights, we update only the weights of a light-weight parameter-efficient finetuning (PEFT) module.
As our PEFT module, we use a DoRA adaptor with parameters: rank $r=16$ and $\alpha=16$. 
We apply the module only to the attention matrices: \path{["q_proj","k_proj","v_proj","o_proj"]}. 
When finetuning the SM versions, we only use the \emph{two pass approach} and activate SM with $p_{\text{sm}} = 0.5$. We didn't see as much of an effect as language modeling with varying this value.
We use an AdamW optimizer with cosine scheduling, a 0.03 warmup ratio and a max gradient norm of 7.0. 
The learning rates are capped at
$\eta_{\text{bb}}=$1e-5 and $\eta_{\text{sm}}=$1e-2 for the DoRA and the SM parameters, respectively.
The finetuning is performed with an effective batch size of 8 on one A100 40GB GPU and takes about 71 hours for 33.5k gradient steps.

\paragraph{Training corpus}
\label{training_set}
For the training corpus, we aimed to use the same datasets that were used by the Dream models~\citep{ye_dream_2025, xie_dreamcoder_2025} in the SFT phase of their Instruct model training. 
\begin{itemize}
    \item \textbf{Dream-7B} The authors report instruction-tuning with \emph{...1.8M instruction-response pairs from Tulu 3~\citep{tulu3_2025} and SmolLM2~\citep{smol_2025}...} 
    We use the same mix. 
    For the Tulu 3 data, we use \path{allenai/tulu-3-sft-mixture}, consisting of 939,000 pairs in their \emph{training} set. Since there is no validation set, we hold out a random 1\% of these pairs for validation.
    For the SmolLM2 data, we use \path{HuggingFaceTB/smoltalk}, consisting of both a training set with 1.04M pairs and a test set with 54.9k pairs. We used the defined \emph{test} set for validation. 
    These datasets are the concatenated and shuffled to make up our training corpus.

    \item \textbf{Dream-Coder-7B} For the Coder model, \citet{xie_dreamcoder_2025} specify explicitly that they use \path{inclusionAI/Ling-Coder-SFT}~\citep{codefuse2025samplemattersleveragingmixtureofexperts} for their SFT training. We use the same dataset. This dataset consists of 4.48M pairs. We hold out a random 1\% at the beginning to be used for validation. 
\end{itemize}

\paragraph{Preprocessing} The preprocessing of these question-answer pairs is executed as follows:
\begin{enumerate}
    \item \textbf{Max train/validation size.} We first sample 300,000 datapoints to use for training and 500 to use for evaluation. 
    
    \item \textbf{Context and response splitting.} All datapoints contain a sequence of \emph{user} and \emph{assistant} messages. We consider the last assistant message to be the \emph{response} and all other previous messages to make up the \emph{context}. We apply the respective models' chat template to the messages before tokenization.

    \item \textbf{Filtering the dataset.} We conduct dataset filtering on three different attributes of the data: \textbf{(1)} inputs $> 2048$ tokens; \textbf{(2)} responses $< 5$ tokens; \textbf{(3)} any examples with tool-calling (i.e., using $<$tool\_call$>$ and $<$/tool\_call$>$). 
    After filtering, we are typically left with just less than 270,000 context-response pairs. 
    While this varies with our sampling seed, the total number of tokens in our training set is often just under 200 million.
    These 200 million tokens consist of a fifty-fifty split between prompt and response tokens.
    
    \item \textbf{Padding with $<$eos$>$ tokens.} 
    After finetuning, we want the diffusion model to retain its ability to decide when to end its generation process. 
    In MDLMs, this is typically done by predicting $<$eos$>$ tokens for all the end positions that the model does not want to use. 
    In order to ensure this is a part of the training process we add $n_\text{end} \sim$ Uniform$(0,50)$ $<$eos$>$ tokens to the end of each training sample.
    %
    %
    The amount of padding that is added to the training set varies for each sample. 

    \item \textbf{Partially masking the responses for training.} We only apply masks to the response tokens during the training process. This is performed in the following way: 
    First, as Algorithm~\ref{alg:training} describes, a mask probability value $t \sim \text{Uniform}(b_l,b_h)$ is sampled. 
    This value $t$ is then used as the masking probability: a fraction $t$ of the \emph{response} tokens are masked for the loss calculation.
    \citet{arriola_block_2024} found that sampling extremely low/high $t$ can lead to high variance in the gradient norms, making training quite difficult. For this reason, we use a clipped noise schedule of $(b_l, b_h) = (0.2,0.8)$.
    The masks are sampled uniquely for each sample. 
\end{enumerate}

\paragraph{SM-feedback parameter initialization}
\label{app:initialization}
As mentioned in Section~\ref{sec:confidence_scaling}, we introduced a mapping from the entropy to the amount of feedback:
\begin{align}
    \lambda^l (\mathbf{p}^l_{t-1}) = \omega_s\cdot \sigma \Big(\omega_a\big(-H(\mathbf{p}^l_{t-1})-\omega_b\big)\Big). 
\end{align}
These parameters: $\omega_a, \omega_b$ and $\omega_s$ are learnable during the training process. 
We ensure that the scale value, $s$, is initialized close to 0 at the start (slightly larger due to the sigmoid reparameterization discussed in Appendix~\ref{app:parameterization}. 
This ensures that the model only adds SM-feedback if it learns that this is optimal via the finetuning process.
For the steepness and center of the sigmoid, we conduct a small statistical analysis of the expected entropy distribution.
Although the theoretical range of $-H(\mathbf{p}^l_{t-1})$ is $[-\log(|\mathcal{V}|), 0]$, in practice we observed that 95\% of values were above $\text{LB} \approx -1.5$.  
For this reason, we initialize the center of the sigmoid at $b = \text{LB}/2$ and choose $a = -10/\text{LB}$. Effectively, this normalizes the negative entropy to be between $[0,1]$ before applying a sigmoid of $a'=10, b'=0.5$.
We found that the learning rate for the hyperparameters needed to be set much higher than for the DoRA adaptor in order for them to be able to traverse the entire range of options. 
The rate was set to $\eta_{\text{sm}}=$1e-2 for the training process. 

\paragraph{Fast-dLLM Throughput Experiment}
\label{app:fast_dllm}

Since Fast-dLLM~\citep{wu_fastdllm_2026} performs blockwise decoding, there is a noticeable correlation between the block lengths and the inference time. Longer blocks allow much more parallelism in decoding and lead to a faster response, while shorter block lengths tend to drift the model towards more autoregressive generation. 
For our experiment, we used block lengths of 16, 32, 48, 96, 128, 192, 256, and a total generation length of 768. These led to fractional lengths of 1/48, 1/24, 1/16, 1/8, 1/6, 1/4, 1/3. We then calculated the throughput as the total tokens generated (for all questions) divided by the total decoding time. The results are shown in Figure~\ref{fig:fast_dllm}.

\newpage
\section{More Results and Ablations}

\subsection{Language Modeling: OWT Validation Perplexity}\label{app:val-perplexity}
In this appendix, we benchmark the validation perplexity of SM against state-of-the-art methods. 
We note that, in contrast to standard baselines, SM necessitates two model passes for perplexity evaluation. 
As demonstrated in Table~\ref{tab:owt-perplexity}, our method achieves superior performance compared to all other diffusion models in the iso-update regime, including a concurrent work (LDDM-M;~\citealt{jo_loopholing_2026}). 
Furthermore, even under the stricter iso-compute constraint, SM outperforms MDLM and SEDD and remains competitive with GIDD+. 
\begin{table}[h]
\caption{
Validation perplexity on OWT.
$^\dagger$Results for AR and SEDD were taken from ~\citep{sahoo_diffusion_2025}. 
}
\centering
\label{tab:owt-perplexity}
\resizebox{\linewidth}{!}{%
\begin{tabular}{lrrrr}
\toprule
                          & Gradient updates & Forward passes & Training tokens & PPL         \\
\cmidrule(r){2-4}\cmidrule(r){5-5}
AR$^\dagger$                        & 0.5M             & 0.5M           & 262B             & 17.54       \\
SEDD$^\dagger$~\citep{lou_discrete_2024}                     & 1M               & 1M             & 262B             & $\leq24.10$ \\
MDLM$^\dagger$~\citep{sahoo_simple_2024}                   & 1M               & 1M             & 262B             & $\leq23.21$ \\
GIDD+~\citep{rutte_generalized_2025}                     & 1M               & 1M             & 262B            & $\leq22.29$ \\
LDDM-M~\citep{jo_loopholing_2026} & 1M               & 2M             & 262B            & $\leq21.90$ \\
Our MDLM+SM (iso-compute) & 0.5M             & 1M             & 131B             & $\leq22.36$ \\
Our MDLM+SM (iso-update)  & 1M               & 2M             & 262B             & $\leq\mathbf{21.47}$ \\
\bottomrule
\end{tabular}
}
\end{table}

\subsection{Language Modeling: Generative Perplexity and Entropy}\label{app:entropy}
This section analyzes the generative perplexity and the entropy observed during unconstrained generation.
Table~\ref{tab:scratch-entropy} shows the generative perplexity and the entropy when training MDLM with SM from scratch.
The reported generative perplexity is repeated from Table~\ref{tab:scratch} to facilitate comparison.
As shown, our SM maintains an entropy on par with the binary masking baseline.
For ReMDM unmasking, SM shows its lowest entropy at the 1/2 NFE budget ($5.234$ and $5.217$ for iso-compute and iso-update, respectively). While still being higher than binary masking entropy ($5.209$), this indicates a highly low-diversity output, which likely explains the observed degradation in the MAUVE score, despite the simultaneously low generative perplexity ($11.40$ and $10.85$). 
This suggests that the ReMDM unmasking process might require specific hyperparameter tuning to balance diversity with human-like text generation (MAUVE).

Table~\ref{tab:gen_ppl_cont} and Table~\ref{tab:gen_entropy_cont} respectively show the generative perplexity and entropy for the continuation of pretraining.
Here, too, SM consistently improves the generative perplexity while maintaining the entropy.
\begin{table*}[h!]
\caption{
Generative perplexity ($\downarrow$) and entropy ($\uparrow$) with pretraining from scratch.
We perform unconstrained generation of $L=1024$ tokens using MDLM~\citep{sahoo_simple_2024} with binary masking or our SM. Evaluations are tabulated by varying NFE budgets. For unmasking, we use either the standard or the more recent ReMDM~\citep{wang_remasking_2025}; the highest scores are bolded. \textit{Gain} shows the performance improvement between the SM and the baseline MDLM.
}
\centering
\label{tab:scratch-entropy}
\resizebox{\linewidth}{!}{%
\begin{tabular}{lllllllllllll}
\toprule
 &  & \multirow{2}{*}{Gradient} & \multirow{2}{*}{Forward} & \multicolumn{4}{c}{Generative Perplexity $\downarrow$} & \multicolumn{4}{c}{Entropy $\uparrow$} \\
\cmidrule(r){5-8}\cmidrule(r){9-12}
Unmasking & Feedback & updates  & passes & 1/8 & 1/4 & 1/2 & 1/1 & 1/8 & 1/4 & 1/2 & 1/1 \\
\cmidrule(r){1-4}\cmidrule(r){5-8}\cmidrule(r){9-12}
\multirow{5}{*}{Standard} 
& {Binary} & {1M} & {1M} & {$60.02$} & {$54.95$} & {$52.36$} & {$50.46$} & {$5.508$} & {$5.482$} & {$5.464$} & {$5.450$}  \\
& {Our SM (iso-compute)} & {0.5M} & {1M} & {${41.08}$} & {${31.97}$} & {${27.36}$} & {${24.63}$} & {$5.496$} & {$5.448$} & {$5.409$} & {$5.374$}    \\
& \quad \textit{{Gain}} & & & \gain{-18.93} & \gain{-22.98} & \gain{-24.99} & \gain{-25.83} \\
& {Our SM (iso-update)} & {1M} & {2M} & {$ \mathbf{39.61}$} & {$\mathbf{30.74}$} & {$\mathbf{26.12}$} & {$\mathbf{23.53}$} & {$5.488$} & {$5.438$} & {$5.398$} & {$5.357$} \\
& \quad {\textit{Gain}} & & & \gain{-20.41} & \gain{-24.21} & \gain{-26.23} & \gain{-26.93} & \\
\cmidrule(r){1-4}\cmidrule(r){5-8}\cmidrule(r){9-12}
\multirow{5}{*}{ReMDM} 
& {Binary} & {1M} & {1M} & {$42.53$} & {$31.05$} & {$21.75$} & {$28.62$}  & {$5.424$} & {$5.336$} & {$5.209$} & {$5.368$}     \\
& {Our SM (iso-compute)} & {0.5M} & {1M} & {${29.90}$} & {${18.08}$} & {${11.40}$} & {${17.29}$} & {$5.424$} & {$5.334$} & {$5.234$} & {$5.349$}  \\
& \quad \textit{{Gain}} & & & \gain{-12.63} & \gain{-12.97} & \gain{-10.35} & \gain{-11.33} \\
& {Our SM (iso-update)} & {1M} & {2M} & {$\mathbf{29.62}$} & {$\mathbf{17.58}$} & {$\mathbf{10.85}$} & {$\mathbf{16.72}$} & {$5.416$} & {$5.323$} & {$5.217$} & {$5.338$} \\
& \quad {\textit{Gain}} & & & \gain{-12.91} & \gain{-13.48} & \gain{-10.90} & \gain{-11.90} \\
\cmidrule(r){1-4}\cmidrule(r){5-8}\cmidrule(r){9-12}
\multicolumn{4}{c}{AR ($T=1024$)} & \multicolumn{4}{c}{12.1} & \multicolumn{4}{c}{5.22}\\ 
\cmidrule(r){1-4}\cmidrule(r){5-8}\cmidrule(r){9-12}
\multicolumn{4}{c}{Data} & \multicolumn{4}{c}{14.8} & \multicolumn{4}{c}{5.44}\\ 
\bottomrule
\end{tabular}
}
\end{table*}

\begin{table*}[h!]
\caption{
{
Generative perplexity ($\downarrow$) with pretraining continuation. We perform unconstrained generation of $L=1024$ tokens using MDLM~\citep{sahoo_simple_2024} with binary masking or our SM. For unmasking, we use either standard or ReMDM~\citep{wang_remasking_2025}; the highest scores are bolded. \textit{Gain} shows the performance improvement between the SM and the binary MDLM with pretraining continuation.
}
}
\centering
\label{tab:gen_ppl_cont}
\resizebox{\linewidth}{!}{%
\begin{tabular}{lllllll}
\toprule
 &  & \multirow{2}{*}{Pretraining} & \multicolumn{4}{c}{Number of function evaluations (NFEs)} \\
\cmidrule(r){4-7}
Unmasking & Feedback & steps  & 1/8 & 1/4 & 1/2 & 1/1 \\
\cmidrule(r){1-3}\cmidrule(r){4-7}
\multirow{5}{*}{Standard} 
& {Binary} & {1M} & {$60.02$} & {$54.95$} & {$52.36$} & {$50.46$}      \\
& {Binary} & {1M+100k} & {$59.99_{(\pm0.68)}$} & {$54.71_{(\pm0.55)}$} & {$52.15_{(\pm0.69)}$} & {$50.63_{(\pm0.72)}$}  \\
& {Our SM (iso-compute)} & {1M+50k} & {$51.10_{(\pm0.89)}$} & {$43.25_{(\pm0.80)}$} & {$39.44_{(\pm0.74)}$} & {$37.46_{(\pm0.80)}$}  \\
& \quad {\textit{Gain}} & & \gain{-8.89} & \gain{-11.46} & \gain{-12.71} & \gain{-13.18} \\
& {Our SM (iso-update)} & {1M+100k} & {$\mathbf{50.99}_{(\pm0.41)}$} & {$\mathbf{42.75}_{(\pm0.39)}$} & {$\mathbf{38.56}_{(\pm0.43)}$} & {$\mathbf{35.81}_{(\pm1.24)}$}
   \\
& \quad {\textit{Gain}} & & \gain{-9.00} & \gain{-11.97} & \gain{-13.59} & \gain{-14.82} \\
\cmidrule(r){1-3}\cmidrule(r){4-7}
\multirow{5}{*}{ReMDM} 
& {Binary} & {1M} & {$42.53$} & {$31.05$} & {$21.75$} & {$28.62$}    \\
& {Binary} & {1M+100k} & {$42.85_{(\pm0.68)}$} & {$31.07_{(\pm0.39)}$} & {$21.74_{(\pm0.38)}$} & {$28.65_{(\pm 0.33)}$}  \\
& {Our SM (iso-compute)} & {1M+50k} & {$39.61_{(\pm0.87)}$} & {$26.29_{(\pm0.73)}$} & {$17.65_{(\pm0.47)}$} & {$22.47_{(\pm0.36)}$}  \\
& \quad {\textit{Gain}} & & \gain{-3.24} & \gain{-4.78} & \gain{-4.08} & \gain{-6.18} \\
& {Our SM (iso-update)} & {1M+100k} & {$\mathbf{39.52}_{(\pm0.33)}$} & {$\mathbf{25.93}_{(\pm0.22)}$} & {$\mathbf{17.23}_{(\pm0.16)}$} & {$\mathbf{22.10}_{(\pm0.21)}$}  \\
& \quad {\textit{Gain}} & & \gain{-3.33} & \gain{-5.14} & \gain{-4.51} & \gain{-6.54} \\
\bottomrule
\end{tabular}
}
\end{table*}

\begin{table*}[h!]
\caption{
Entropy ($\uparrow$) with pretraining continuation. We perform unconstrained generation of $L=1024$ tokens using MDLM~\citep{sahoo_simple_2024} with binary masking or our SM. For unmasking, we use either standard or ReMDM~\citep{wang_remasking_2025}. 
}
\centering
\label{tab:gen_entropy_cont}
\resizebox{\linewidth}{!}{%
\begin{tabular}{lllllll}
\toprule
 &  & \multirow{2}{*}{Pretraining} & \multicolumn{4}{c}{Number of function evaluations (NFEs)} \\
\cmidrule(r){4-7}
Unmasking & Feedback & steps  & 1/8 & 1/4 & 1/2 & 1/1 \\
\cmidrule(r){1-3}\cmidrule(r){4-7}
\multirow{5}{*}{Standard} 
& {Binary} & {1M} & {$5.508$} & {$5.482$} & {$5.464$} & {$5.450$}      \\
& {Binary} & {1M+100k} & {$5.503_{(\pm0.005)}$} & {$5.477_{(\pm0.006)}$} & {$5.458_{(\pm0.005)}$} & {$5.447_{(\pm0.007)}$}  \\
& {Our SM (iso-compute)} & {1M+50k} & {$5.534_{(\pm0.004)}$} & {$5.503_{(\pm0.004)}$} & {$5.480_{(\pm0.003)}$} & {$5.467_{(\pm0.007)}$}  \\
& {Our SM (iso-update)} & {1M+100k} & {${5.542}_{(\pm0.003)}$} & {${5.509}_{(\pm0.003)}$} & {${5.485}_{(\pm0.004)}$} & {${5.453}_{(\pm0.036)}$}
   \\
\cmidrule(r){1-3}\cmidrule(r){4-7}
\multirow{5}{*}{ReMDM} 
& {Binary} & {1M} & {$5.424$} & {$5.336$} & {$5.209$} & {$5.368$}    \\
& {Binary} & {1M+100k} & {$5.423_{(\pm0.007)}$} & {$5.333_{(\pm0.007)}$} & {$5.200_{(\pm0.005)}$} & {$5.361_{(\pm0.005)}$}  \\
& {Our SM (iso-compute)} & {1M+50k} & {$5.476_{(\pm0.006)}$} & {$5.396_{(\pm0.008)}$} & {$5.302_{(\pm0.005)}$} & {$5.410_{(\pm0.004)}$}  \\
& {Our SM (iso-update)} & {1M+100k} & {${5.482}_{(\pm0.005)}$} & {${5.404}_{(\pm0.004)}$} & {${5.312}_{(\pm0.003)}$} & {${5.416}_{(\pm0.005)}$}  \\
\bottomrule
\end{tabular}
}
\end{table*}

\subsection{Language Modeling: Training Share, Top-k, Softmax, and Gradient Updates}\label{app:ablation-lm}
\begin{figure}[h!]
    \centering 
    \begin{subfigure}{0.48\textwidth}
        \centering
        \includegraphics[width=\textwidth]{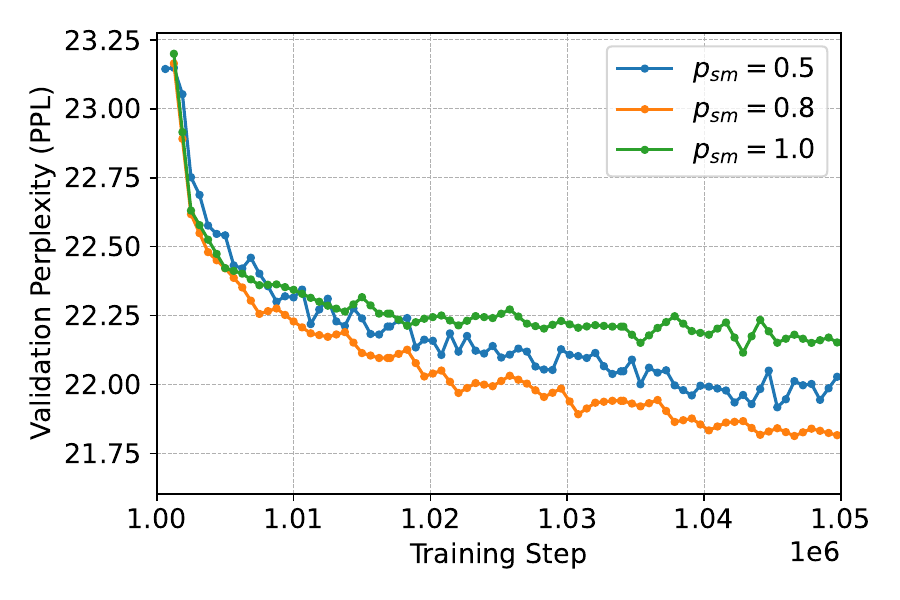}
        \caption{SM training share}
        \label{fig:ablation_lm_p}
    \end{subfigure}
    \hfill 
    \begin{subfigure}{0.48\textwidth}
        \centering
        \includegraphics[width=\textwidth]{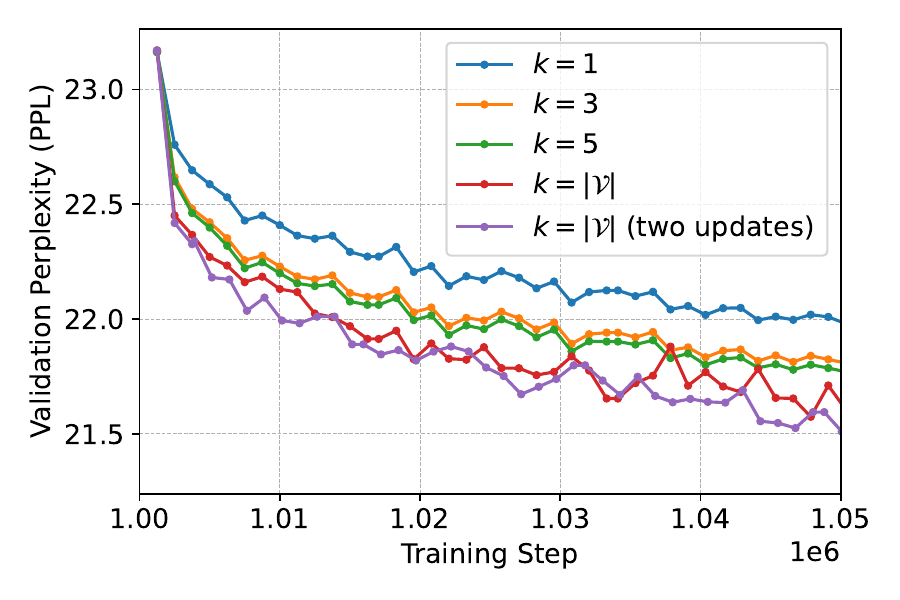}
        \caption{SM feedback and gradient updates}
        \label{fig:ablation_lm_k}
    \end{subfigure}
    \caption{Ablation study language modeling on OWT. Default SM parameters are $p_{sm}=0.8$ and $k=3$.}
    \label{fig:ablation_lm}
\end{figure}

This appendix ablates the SM training probability ($p_{sm}$) and the top-$k$ value in the language modeling experiments.
Recall that $k$ is the number of predicted tokens---generated from the previous step---that will be combined with the mask token.
We start with a default configuration of $p_{sm}=0.8$ and $k=3$ and vary each parameter separately.
As shown in Figure~\ref{fig:ablation_lm_p}, increasing the SM training probability from 0.5 to 0.8 improves the performance.
This is likely due to the model learning to leverage continuous feedback more effectively, as a higher probability means it is exposed to the soft-masking mechanism more frequently during training.
Increasing the SM training probability to 1 is detrimental, as the model loses its ability to handle binary masking, which is essential for the initial decoding steps.

Figure~\ref{fig:ablation_lm_k} shows that the perplexity improves when increasing $k$ from 1 to 5, with only a marginal gain observed between $k=3$ and $k=5$.
Beyond the top-$k$ study, we investigated an alternative SM mechanism using a trainable softmax temperature. 
This method calculates probabilities using a learned temperature applied across the entire vocabulary, $\mathcal{V}$, letting the model scale the range of its own feedback. 
In Figure~\ref{fig:ablation_lm_k}, we denote this ablation $k=|\mathcal{V}|$ to signify that it considers the entire vocabulary. 
As shown, this method can slightly improve the perplexity.
Crucially, the softmax mechanism is differentiable (in contrast to the non-differentiable top-$k$ selection), which allows for the backpropagation of gradients to the first model pass. 
However, we did not observe any further improvements in perplexity when applying the gradients on both forward passes (see the ``$k=|\mathcal{V}|$ (two updates)'').

Overall, we decided to use the top-$k=3$ option as the final configuration. The softmax approach, while theoretically interesting, comes with increased compute and memory demands (due to storing $|\mathcal{V}|$-dimensional vectors) and did not show a conclusive benefit in downstream coding performance (see Appendix~\ref{app:top-k}).

\subsection{Language Modeling: Inference Speed}\label{app:speed}

This appendix analyzes the inference time for both SM and binary masking when unconditionally generating samples using $L=1024$ diffusion steps on an NVIDIA A100 GPU.
We use the inference script provided by ReMDM~\citep{wang_remasking_2025}.
The setup uses ancillary sampling, where the backbone's forward pass is not called if the predictive logits have not changed in the previous iteration (i.e., caching).
SM is configured with $k=3$.
We measure the time using Python's cProfile.

\begin{figure} [h!]
    \centering
    \includegraphics[width=0.6\linewidth]{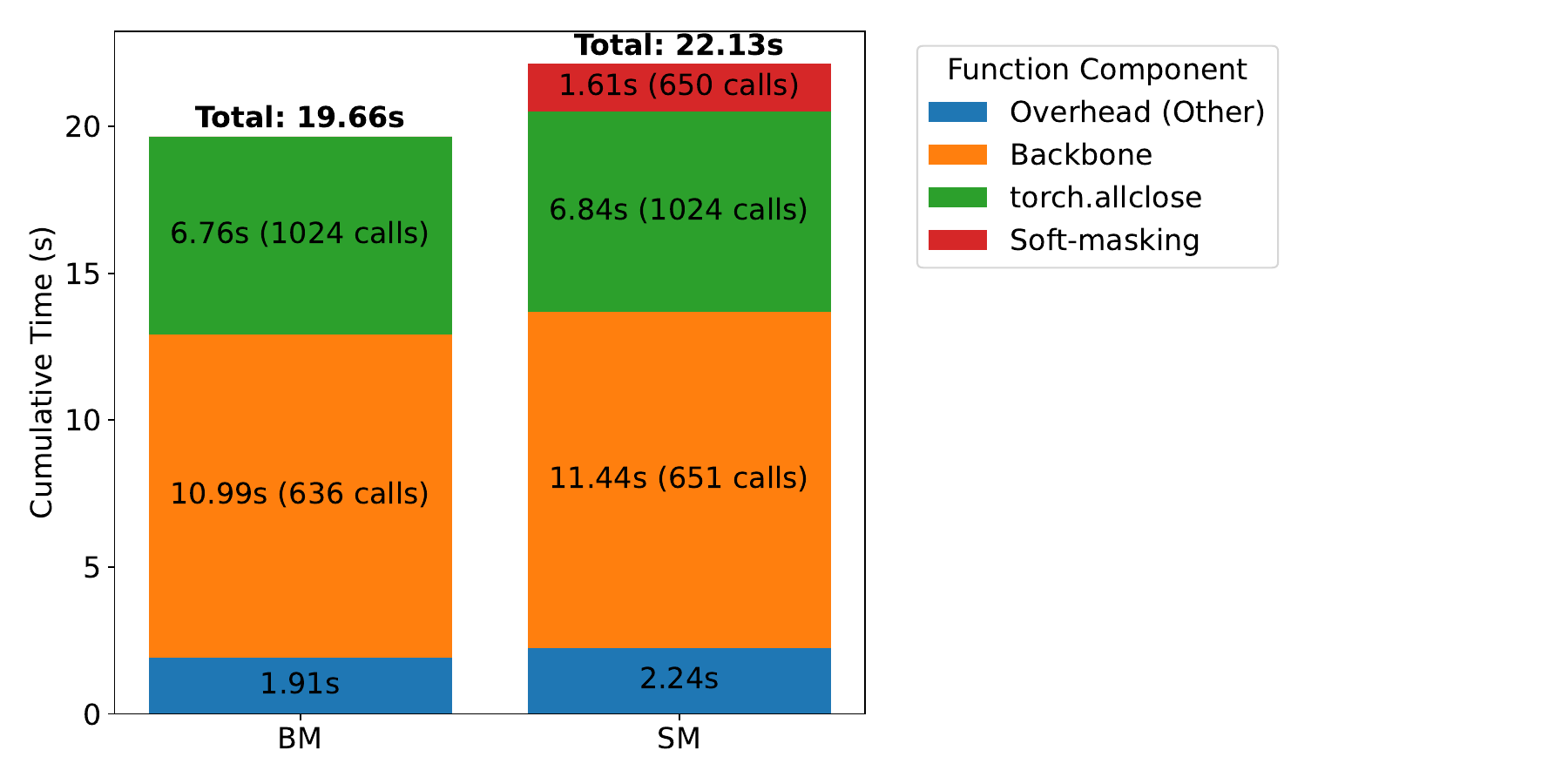}
    \caption{Cumulative time for unconditionally generating L=1024 tokens on an NVIDIA A100 GPU using standard unmasking.}
    \label{fig:lm-inference-time}
\end{figure}

Figure~\ref{fig:lm-inference-time} shows that SM yields a small overhead of 12.5\% (22.26\,s vs. 19.78\,s).
One can see that checking the activation change (`torch.allclose`) yields a major overhead in both configurations.
Diving a bit deeper, one can notice that the MDLM with SM calls the backbone more often than binary masking (651 vs. 636 calls).
This is likely because the more detailed input representation from SM causes the state to change more frequently, resulting in fewer cache hits.
Besides, SM slightly increases the complexity on two fronts.
First, computing the SM distribution requires a total of 1.61\,s (0.00248\,s per call), which is dominated by the top-k computation.
Moreover, the sparse embedding (a weighted sum of $k+1$ tokens) is slightly more complex than the standard embedding (a single token lookup), increasing the per-call time for the backbone by 1.7\% (from 0.01728\,s to 0.01757\,s per call).
In summary, the 2.48\,s total overhead from SM is primarily composed of the 1.61\,s for the SM calculation, with the remaining time due to a 1.7\% increase in per-call backbone cost and a 2.4\% increase in the total number of backbone calls.
Despite the small overhead, the generation quality is significantly increased, as can be seen in Table~\ref{tab:mauve}.

\subsection{Language Modeling: SM Visualization}\label{app:speed}
\begin{figure} [h!]
    \centering
    \includegraphics[width=0.8\linewidth]{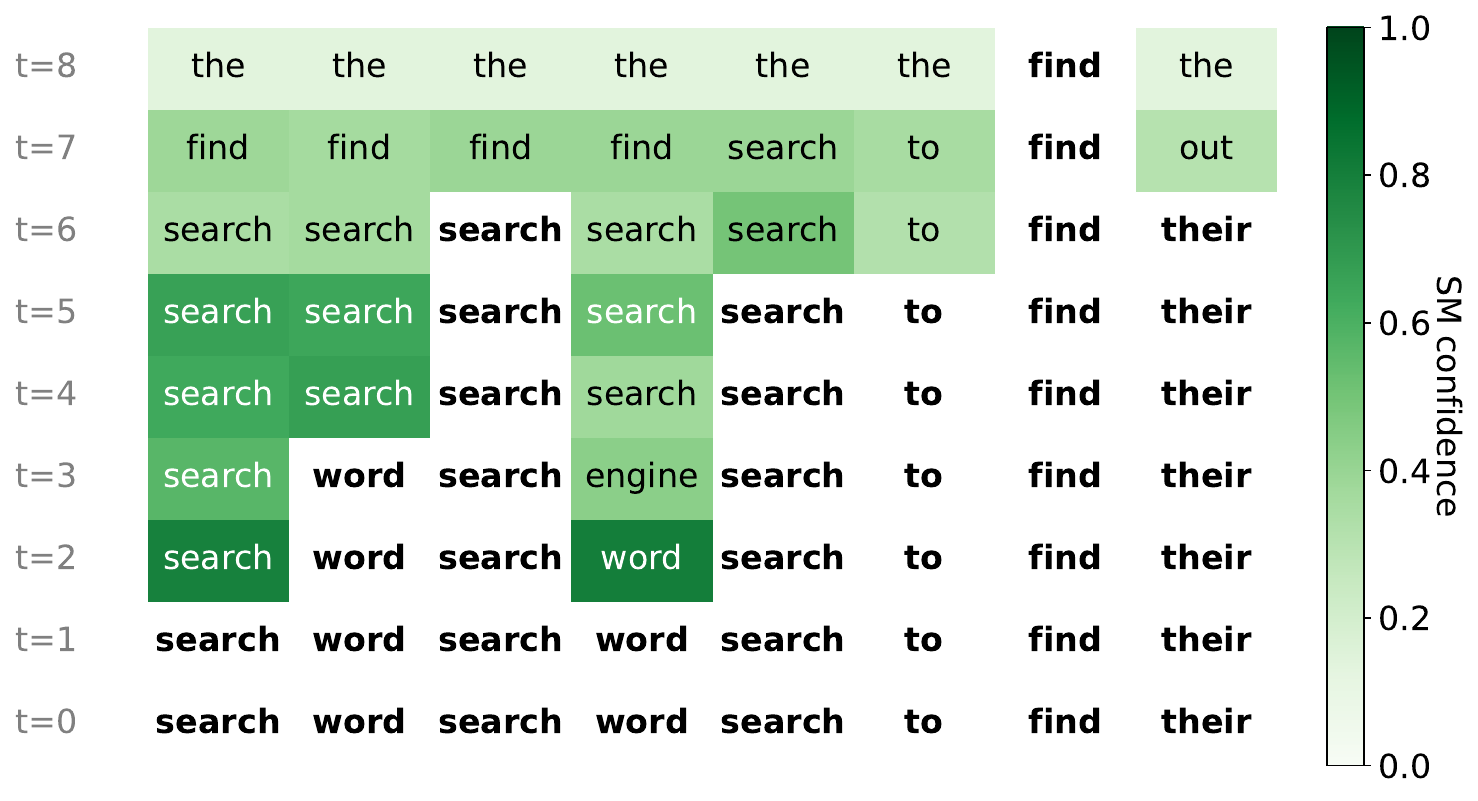}
    \caption{Unconstrained generation trajectory of $L=8$ tokens over $T=8$ steps using an MDLM with Soft-Masking (trained from scratch, iso-compute, 500k steps). Green-shaded cells indicate masked tokens where SM is active; color intensity corresponds to the SM confidence, with darker green indicating higher certainty. The text inside these cells displays the current top-1 predicted token. Bold, unshaded text represents tokens that have been unmasked (sampled) and fixed.
    }
    \label{fig:lm-trajectory}
\end{figure}

This appendix provides further insight into SM's decoding dynamics. 
As an illustrative example, we analyze an unconstrained generation trajectory of $L=8$ tokens over $T=8$ steps. 
Note that we use a standard sampling schedule without caching or confidence-based unmasking heuristics. 
As shown in Figure~\ref{fig:lm-trajectory}, SM's confidence (indicated by color intensity) generally increases over the denoising steps. 
However, high SM confidence does not strictly dictate the final outcome: the top-1 predicted token within the SM evolves over time, and the final sampled token can differ from the highly confident SM prediction of the previous step (e.g., the transition from "search" to "word" at $t=3$ for the second token).

\subsection{Mathematical Reasoning}\label{app:math_reasoning}

We also perform experiments on mathematical reasoning tasks GSM8k~\citep{cobbe_training_2021} and Math-500~\citep{lightman2024math}. These evaluations are shown in Table~\ref{tab:downstream_math}. Both evaluations were performed with \path{lm-evaluation-harness} in a zero-shot setting. The max generation length for GSM8k was 256, and for Math-500 it was set to 512. These values were the same as the default ones used by Dream-7B in their original evaluation scripts. Since the MATH tasks are more difficult, the model often needs more generation space to come to the final answer.
\begin{table}[h!]
\caption{
Accuracy (\%) on math tasks. 
Evaluations are displayed under varying computational NFE budgets.
We finetune the models with 5 seeds and report the mean accuracy ($\pm$ standard deviation). SM is configured with $k$=3.
\textit{Gain} shows the comparison between the SM model and the finetuned baseline. 
}
\label{tab:downstream_math}
\centering
\resizebox{0.5\linewidth}{!}{%
\begin{tabular}{clcll}
\toprule
\multicolumn{1}{l}{}                                 &          & \multicolumn{1}{l}{}    & \multicolumn{2}{c}{Dream-7B} \\
\cmidrule(r){4-5}
\begin{tabular}[c]{@{}c@{}}NFE\\ budget\end{tabular} & Feedback & \begin{tabular}[c]{@{}c@{}}FT\\ steps\end{tabular} & GSM8k    & Math-500  \\
\cmidrule(r){1-3}\cmidrule(r){4-5}
\multirow{4}{*}{1/4}                                 & Binary   & -      &         $57.8$&          $14.2$\\
                                                     & Binary   & 33.5k     &         $59.5_{(\pm 1.1)}$&          $17.3_{(\pm 1.2)}$\\
                                                     & Our SM       & 33.5k  &         $62.3_{(\pm 2.3)}$&          $19.8_{(\pm 2.1)}$\\
                                                     & \quad \textit{Gain}     &                      &         \gain{+2.8}&    \gain{+2.5}\\
                                                     \cmidrule(r){1-3}\cmidrule(r){4-5}
\multirow{4}{*}{1/2}                                 & Binary   & -   &         $76.0$&          $36.6$\\
                                                     & Binary   & 33.5k  &         $76.5_{(\pm 1.6)}$&          $36.3_{(\pm 0.8)}$\\
                                                     & Our SM       & 33.5k   &         $79.4_{(\pm 0.5)}$&          $38.8_{(\pm 1.8)}$\\
                                                     & \quad \textit{Gain}     &   &         \gain{+2.9}&    \gain{+2.5}\\
                                                     \cmidrule(r){1-3}\cmidrule(r){4-5}
\multirow{4}{*}{1/1}                                 & Binary   & -    &         $82.0$&          $45.6$\\
                                                     & Binary   & 33.5k    &         $82.9_{(\pm 1.0)}$&          $42.7_{(\pm 1.4)}$\\
                                                     & Our SM       & 33.5k   &         $84.0_{(\pm 0.7)}$&          $41.4_{(\pm 1.2)}$\\
                                                     & \quad \textit{Gain}     &         &         \gain{+1.1}&    \textit{-1.3}\\
\bottomrule
\end{tabular}
}
\end{table}

\subsection{Coding: Iso-Compute Models}\label{app:coding-isocompute}
Table~\ref{tab:downstream-isocompute} shows SM's performance in the iso-compute training setup. 
Note that in this more restricted setup, the model does only see half of the data, as there is no data repetition in finetuning. 
While the variance across seeds increased slightly, we still observe consistent gains with SM, primarily in low NFE budgets. 
\begin{table}[h!]
\caption{
Accuracy (\%) on coding tasks. SM has been finetuned in the iso-compute training setting. 
Evaluations are tabulated by varying NFE budgets.
We finetune the models with 5 seeds and report the mean accuracy ($\pm$ standard deviation). SM is configured with $k$=1.
\textit{Gain} shows the comparison between the SM model and the finetuned baseline. The best performing model is marked in bold.
}
\label{tab:downstream-isocompute}
\resizebox{\linewidth}{!}{%
\begin{tabular}{clcllllllll}
\toprule
\multicolumn{1}{l}{}                                 &          & \multicolumn{1}{l}{}                                       & \multicolumn{4}{c}{Dream-Coder-7B (instruct)} & \multicolumn{4}{c}{Dream-7B (instruct) } \\
\cmidrule(r){4-7}\cmidrule(r){8-11}
\begin{tabular}[c]{@{}c@{}}NFE\\ budget\end{tabular} & Feedback & \begin{tabular}[c]{@{}c@{}}FT\\ steps\end{tabular} & HumanEval    & HumanEval+    & MBPP    & MBPP+   & HumanEval     & HumanEval+     & MBPP    & MBPP+    \\
\cmidrule(r){1-3}\cmidrule(r){8-11}\cmidrule(r){4-7}\cmidrule(r){8-11}
\multirow{4}{*}{1/4}                                 & Binary   & -                                                          &  $25.0$&        $25.0$&         $27.4$&         {$29.4$}&        $18.9$&         $17.1$&         $26.6$&          {$30.2$}\\
                                                     & Binary   & 33.5k                                                      &       $28.5_{(\pm 1.3)}$&        $27.7_{(\pm 1.8)}$&         $25.9_{(\pm 1.5)}$&         {$24.6_{(\pm 1.7)}$}&        $19.0_{(\pm1.7)}$&         $15.9_{(\pm 2.8)}$&         $27.0_{(\pm 1.6)}$&          {$29.2_{(\pm 1.5)}$}\\
                                                     & {Our SM}       & {16.75k}                                                    &       {$\mathbf{30.4}_{(\pm 1.9)}$}&        {$\mathbf{29.0}_{(\pm 1.3)}$}&         {$\mathbf{27.8}_{(\pm 1.9)}$}&         {$\mathbf{28.8}_{(\pm 3.7)}$}&        {$\mathbf{24.8}_{(\pm 5.7)}$}&         {$\mathbf{22.4}_{(\pm 4.7)}$}&         {$\mathbf{28.2}_{(\pm 1.9)}$}&          {$\mathbf{36.1}_{(\pm 2.0)}$}\\
                                                     & \quad \textit{Gain}     &                                                            &   \gain{+1.9}&        \gain{+1.3}&         \gain{+1.9}&         \gain{+4.3}&        \gain{+5.8}&         \gain{+6.5}&        \gain{+1.2} & \gain{+6.9}   \\
                                                     \cmidrule(r){1-3}\cmidrule(r){8-11}\cmidrule(r){4-7}\cmidrule(r){8-11}
\multirow{4}{*}{1/2}                                 & Binary   & -                                                          &  $54.9$&        $50.6$&         $51.6$&         {$51.3$}&        $31.1$&         $29.3$&         $42.8$&          {$45.8$}\\
                                                     & Binary   & 33.5k                                                      &       $53.8_{(\pm 1.4)}$&        $49.3_{(\pm 1.6)}$&         $\mathbf{49.8}_{(\pm 0.9)}$&         {$53.2_{(\pm 1.5)}$}&        $33.0_{(\pm3.0)}$&         $29.5_{(\pm3.4)}$&         $43.1_{(\pm0.4)}$&          {$39.6_{(\pm2.7)}$}\\
                                                     & {Our SM}       & {16.75k}                                                      &       {$\mathbf{59.4}_{(\pm2.7)}$}&        {$\mathbf{54.3}_{(\pm 1.9)}$}&         {$\mathbf{49.8}_{(\pm 0.5)}$}&         {$\mathbf{55.2}_{(\pm 1.7)}$}&        {$\mathbf{38.5}_{(\pm2.7 )}$}&         {$\mathbf{33.9}_{(\pm2.8)}$}&         {$\mathbf{44.3}_{(\pm2.4 )}$}&          {$\mathbf{53.6}_{(\pm2.8)}$}\\
                                                     & \quad \textit{Gain}     &                                                            &   \gain{+5.6}&        \gain{+5.0}&         \gain{0.0}&         \gain{+2.1}&        \gain{+5.5}&    \gain{+4.4}  &   \gain{+1.2}&        \gain{+14.0}\\
                                                     \cmidrule(r){1-3}\cmidrule(r){8-11}\cmidrule(r){4-7}\cmidrule(r){8-11}
\multirow{4}{*}{1/1}                                 & Binary   & -                                                          &  $75.0$&        $69.5$&         $65.8$&         {$70.4$}&        $57.9$&         $53.0$&         $57.8$&          {$63.5$}\\
                                                     & Binary   & 33.5k                                                      &       $75.7_{(\pm 1.7)}$&        $\mathbf{68.9}_{(\pm 2.0)}$&         $65.6_{(\pm 0.8)}$&         {$68.1_{(\pm 1.1)}$}&        $\mathbf{59.5}_{(\pm1.8)}$&         $\mathbf{53.0}_{(\pm1.0)}$&         $\mathbf{58.3}_{(\pm0.1)}$&          {$\mathbf{62.8}_{(\pm0.7)}$}\\
                                                     & {Our SM}       & {16.75k}                                                     &       {$\mathbf{75.9}_{(\pm1.4 )}$}&        {$68.5_{(\pm0.7 )}$}&         {$\mathbf{66.6}_{(\pm 1.2)}$}&         {$\mathbf{68.6}_{(\pm1.2 )}$}&        {$58.0_{(\pm 2.9)}$}&         {$50.7_{(\pm 3.3)}$}&         {$57.8_{(\pm1.2 )}$}&          {$62.2_{(\pm0.8 )}$}\\
                                                     & \quad \textit{Gain}     &                                                            &   \gain{+0.2}&        \textit{-0.4}&         \gain{+1.0}&         \gain{+0.5}&        \textit{-1.5}&         \textit{-2.3}&         \textit{-0.5}&    \textit{-0.6}\\
\bottomrule
\end{tabular}
}
\end{table}

\subsection{Coding: Top-$k$ and Trainable Softmax Temperature}\label{app:top-k}
We also perform ablation tests on $k$. For these tests, we finetune four models, each with the exact same configuration, but with one key exception: we train each of these models with a different $k\text{-value} \in [1,3,5,10]$. 
We use the same seed for all models to ensure the same training data and initial setup.

We also train a \textit{fifth} model with a \textbf{trainable softmax temperature}. 
This method uses probabilities calculated with a learned temperature instead of using the top-k predicted tokens, letting the model scale the range of it's own feedback. 
In the results table, we call this ablation $k=|V|$

The results given in Table~\ref{tab:k_experiment} illustrate a degrading performance with higher $k$ values, with $k=1$ and $k=3$ having the highest average performance. 
However, all $k$ ablations perform better than both of our baselines on average.
This further illustrates the success of our proposed method.

\input{tables/k_experiment}

\subsection{Coding: Time-dependent Masking}\label{app:time}

We explored three methods of time-dependent (TD) feedback. By time-dependence, we mean scaling the amount of SM feedback as a function of the point in the decoding process (i.e. $t$).
The basic assumption here is that, the model may benefit from having more or less feedback at different steps in the diffusion process.
For the following, let $t = T,...,1$ be our current denoising step, with $T$ being the \textit{first} step in the \textit{reverse} process. 
Let $g(\mathbf{p}) = \omega_s\cdot \sigma \Big(\omega_a\big(-H(\mathbf{p}^l_{t-1})-\omega_b\big)\Big)$ be the \emph{default} (non-time-dependent) defined in~\eqref{eq:confidence_scaling}.
\begin{enumerate}
    \item \textbf{No TD.} In this adaptation, we apply no time-dependent feedback modification. SM is applied exactly as described above. 
    \item \textbf{Stepwise TD feedback function.} This time-dependent feedback defines a threshold value $0 \leq  t' \leq T$. At the threshold, the model switches between SM and Binary as follows:
    \begin{align*}
        \textbf{Binary$\rightarrow$SM stepwise TD function: }& \lambda^l (\mathbf{p}^l_{t-1}) = g(\mathbf{p})\cdot \mathbf{1}_{(t \leq t')}\\
        \textbf{SM$\rightarrow$Binary stepwise TD function: }& \lambda^l (\mathbf{p}^l_{t-1}) = g(\mathbf{p})\cdot \mathbf{1}_{(t \geq t')}
    \end{align*}
    \item \textbf{Linear TD feedback function.} For the last formulation, we add a linear time-dependence to the feedback magnitude. Again, this entails a model switch from SM to Binary, or vice versa. The switch happens with a linear transition function:
    \begin{align*}
        \textbf{Binary$\rightarrow$SM linear TD function: }& \lambda^l (\mathbf{p}^l_{t-1}) = \left(1-\frac{t}{T}\right)\left[g(\mathbf{p})\right]\\
        \textbf{SM$\rightarrow$Binary linear TD function: }& \lambda^l (\mathbf{p}^l_{t-1}) = \left(\frac{t}{T}\right)\left[g(\mathbf{p})\right]
    \end{align*}
\end{enumerate}
Note that these expressions are written this way because, in the reverse process, the time index $t$ \textit{decreases} from $T \rightarrow 1$. 
Although we did not explicitly finetune the models to incorporate external time-dependence (TD) in the feedback function, we perform ablation studies on the TD during inference.
\subsubsection{Early vs. Late Stage SM Impact}

We first look at a simple comparison of whether SM is more beneficial at the \textit{early} or \textit{late} stages of denoising.
To test this, we use the \textbf{stepwise feedback function}:
Specifically, we compare the \textbf{SM$\rightarrow$Binary} and the \textbf{Binary$\rightarrow$SM} stepwise TD functions. 
We define each with thresholds set such that $80\%$ of denoising steps use SM and the remaining $20\%$ use Binary:
\begin{enumerate}
    \item \textbf{Binary$\rightarrow$SM stepwise TD feedback with threshold $\mathbf{t'=0.8}$:} SM only during the \textit{last} $80\%$ of the denoising steps.
    
    \item \textbf{SM$\rightarrow$Binary stepwise TD feedback with threshold $\mathbf{t'=0.2}$:} SM only during the \textit{first} $80\%$ of the denoising steps.
\end{enumerate}

The results given in Table~\ref{tab:time_dependence} evaluate both approaches, comparing them with both binary baselines. 
We can clearly see that ablation \textbf{(2)} performs significantly better than ablation \textbf{(1)}, suggesting that SM is necessary during the early denoising steps.

\input{tables/time-dependence}

\subsubsection{When Is SM Most Beneficial?}

After confirming that SM is necessary during the early denoising steps, we can start looking at the exact steps of the denoising where SM is more beneficial.
To further understand this time-dependence, we compare five forms of our $\textbf{SM$\rightarrow$Binary}$ TD feedback: 
\textbf{(1)} no time-dependence (TD); 
\textbf{(2)} linear SM$\rightarrow$Binary TD; 
\textbf{(3)-(5)} SM$\rightarrow$Binary stepwise TD feedback with thresholds of $\mathbf{t'=0.2}$, $\mathbf{t'=0.5}$, and $\mathbf{t'=0.8}$.

The results are given in Table~\ref{tab:time_dependence_early}.
We tabulate with the \textit{mean} of the finetuned binary models to show that all given TD ablations of SM still perform better than the binary feedback.

\input{tables/time-dependence-early}

\subsection{Coding: Learned SM-feedback Parameters}
\label{app:learned_parameters}

\begin{figure}[h!]
     \centering
    \includegraphics[width=\textwidth]{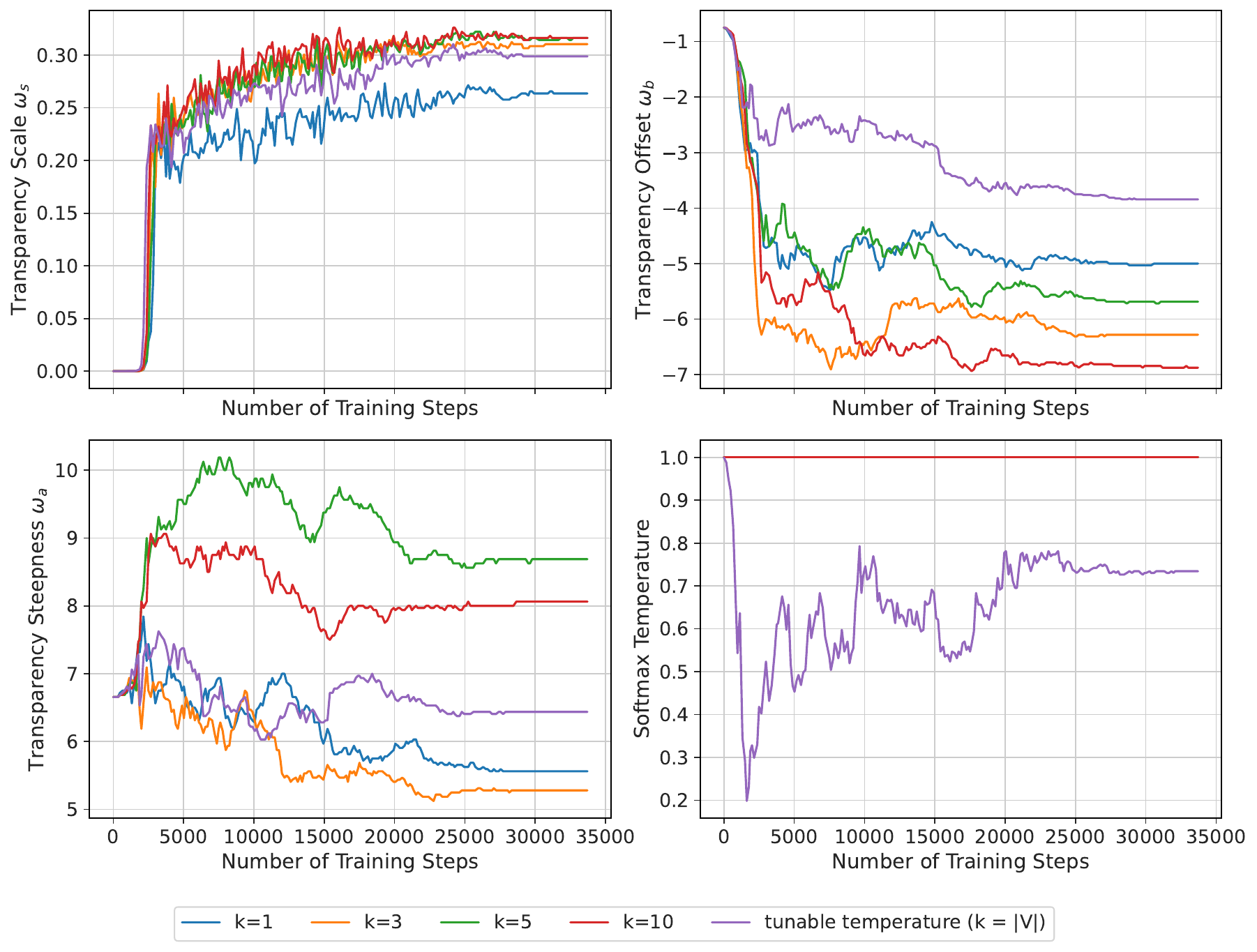}
    \caption{Plots of all four tunable hyperparameters over all of my SM-FT runs and ablations. We show the evolution of the SM-scale (top-left), the SM-offset (top-right), the SM-steepness (bottom-left), and the tunable softmax temperature (bottom-right) (for the $k=|V|$ ablation described in Section~\ref{app:top-k}). 
    }
    \label{fig:learned_th_parameters}
\end{figure}

As we discussed in Section~\ref{sec:confidence_scaling}, we add the SM-feedback parameters: SM-scaling~$=\omega_s$, SM-offset~$=\omega_b$, and SM-steepness~$=\omega_a$ to the computation graph, allowing the optimizer to train these parameters. We also add the softmax temperature for our $k=|\mathcal{V}|$ ablation (Section~\ref{app:top-k}). 
The progression of these parameters throughout the full training of our SM-FT models is illustrated in Figure~\ref{fig:learned_th_parameters}. We find that the learned scale is much lower than in the language modeling. This is likely due to the absence of time conditioning. Since the model has no other cues telling it at what stage it is at in the decoding and/or which tokens are actually masked, it relies on the existence of the mask tokens for this information.

\section{Concurrent Works}\label{app:concurrent}
Concurrently with the development of our SM, the following works also aim to address binary MDLM's weakness in discarding valuable information during the denoising iterations. 

Continuously Augmented Discrete Diffusion (CADD; ~\citealt{zheng_continuously_2026}) models the forward diffusion transition function as a product of a discrete (masking) and a continuous (Gaussian) diffusion process.
This formulation permits a closed-form calculation of the forward process marginals, thereby enabling training with a single forward pass.
In CADD, the input to the denoising model is a superposition of the \mask embedding, the embedding of the current top-1 token estimate, and a noise vector drawn from a normal distribution.
The authors also propose a ``soft" variant that computes a convex combination over the entire vocabulary (our $k=|\mathcal{V}|$). 
However, they do not show any downstream performance on language modeling and coding due to the high computational costs. 
In contrast, our SM covers the entire spectrum between hard (top-1) and soft through top-$k$ superposition, making a beneficial trade-off between generation quality and computational cost (see Appendix~\ref{app:speed}). 
Moreover, while CADD employs a deterministic schedule to define the weighting of the estimated mean vector, our SM utilizes a learnable, confidence-based weighting mechanism.
Table~\ref{tab:cadd} compares the natural language modeling performance of our SM against CADD.
Our SM achieves notably higher MAUVE scores in both iso-compute and iso-update settings. 
A similar trend is observed for the generative perplexity.
Finally, differences in evaluation setups (qwencoder-eval vs. lm-evaluation-harness) and base models (DiffuCoder vs. DreamCoder) render a direct comparison on coding benchmarks difficult.

\begin{table*}[h!]
\caption{
Comparison to concurrent CADD~\citep{zheng_continuously_2026}. 
All methods perform unconstrained generation with sequence length $L=1024$. Evaluations are tabulated by varying NFE budgets. Standard unmasking is used; the highest scores are bolded. 
}
\centering
\label{tab:cadd}
\resizebox{\linewidth}{!}{%
\begin{tabular}{lllllllllllll}
\toprule
 & \multirow{2}{*}{Batch} & \multirow{2}{*}{Gradient} & \multirow{2}{*}{Forward} & \multirow{2}{*}{Loss} & \multicolumn{4}{c}{MAUVE $\uparrow$} & \multicolumn{4}{c}{Generative Perplexity $\downarrow$} \\
\cmidrule(r){6-9}\cmidrule(r){10-13}
 Feedback & size & updates  & tokens & tokens & 1/8 & 1/4 & 1/2 & 1/1 & 1/8 & 1/4 & 1/2 & 1/1 \\
\cmidrule(r){1-5}\cmidrule(r){6-9}\cmidrule(r){10-13}
MDLM & 512 & {1M} & 524B & 262B & {$0.017$} & {$0.025$} & {$0.036$} & {$0.034$}  & {$60.0$} & {$55.0$} & {$52.4$} & {$50.5$}    \\
CADD~\citep{zheng_continuously_2026} & 256 & {2M} & 524B & 262B & $0.017$ & $0.080$ & $0.140$ & $0.240$  &  $55.3$ & $49.6$ & $46.1$ & $44.6$  \\
 {Our SM (iso-compute)} & 512 & {0.5M} & 524B & 131B & {$0.143$} & {$\mathbf{0.417}$} & {${0.498}$} & {${0.596}$} & {$41.1$} & {${32.0}$} & {${27.4}$} & {${24.6}$}  \\
 {Our SM (iso-update)} & 512 & {1M} & 1049B & 262B & {$\mathbf{0.155}$} & {$0.383$} & {$\mathbf{0.535}$} & {$\mathbf{0.602}$} & {$ \mathbf{39.6}$} & {$\mathbf{30.7}$} & {$\mathbf{26.1}$} & {$\mathbf{23.5}$} \\  
\bottomrule
\end{tabular}
}
\end{table*}

\begin{figure}
    \centering
    \includegraphics[width=1.0\linewidth]{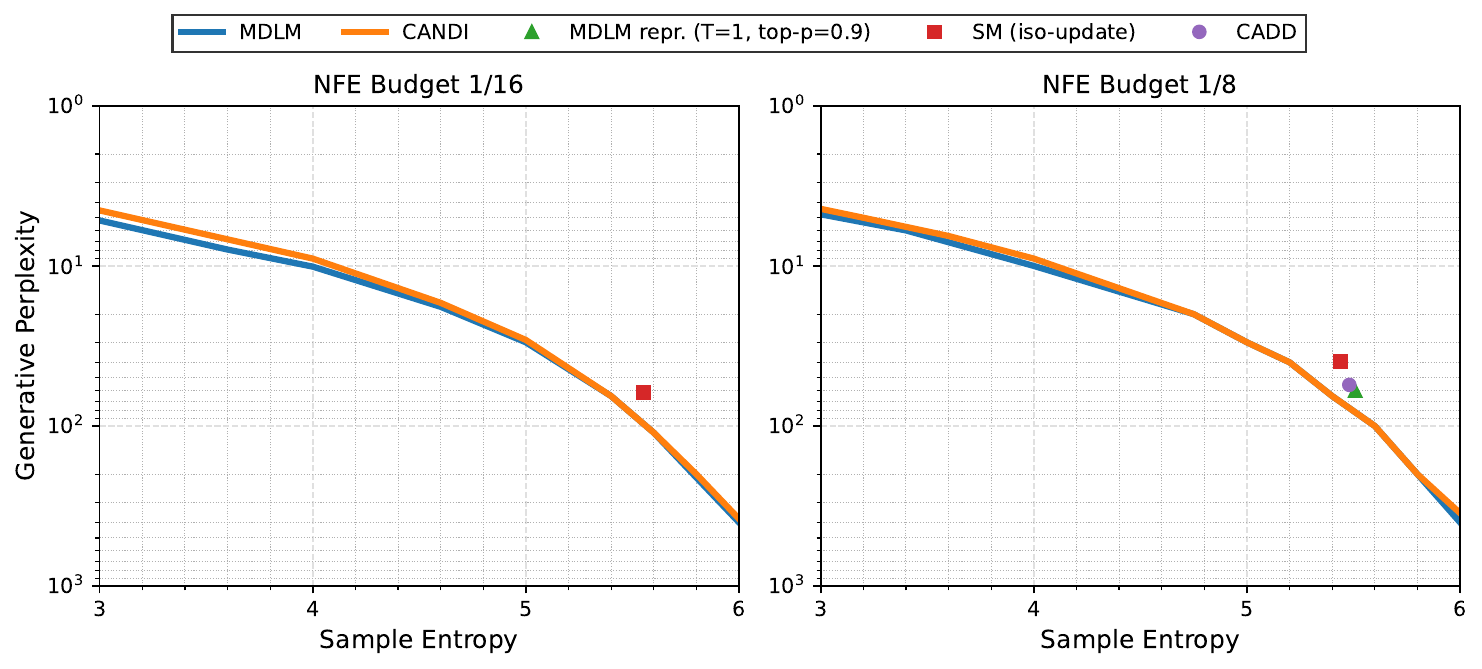}
    \caption{Comparison to CANDI~\citep{pynadath_candi_2025} and CADD~\citep{zheng_continuously_2026} on unconstrained text generation. MDLM and CANDI results were visually extracted from~\citep{pynadath_candi_2025}. Our SM yields the best trade-off between generative perplexity and entropy.}
    \label{fig:candi_comp}
\end{figure}

Continuous and Discrete Diffusion (CANDI; \citealt{pynadath_candi_2025}) analyzes the limitations of continuous diffusion models on discrete text through a framework based on token identifiability.
Similar to CADD, CANDI merges discrete and continuous diffusion via the product of the respective transition functions.
The score function is approximated via a single-step Monte-Carlo estimation, which uses the sampled token embedding.
The denoising model is conditioned on a superposition of the current estimate, Gaussian noise, and a mask token, weighted by a fixed hyperparameter ($\lambda$). 
While allowing for closed-form marginal computation and thus single-pass training, their formulation does not involve any confidence-based weighting.
Experimentally, the authors find that the unconstrained text generation is highly dependent on their sampling temperature. 
They show that CANDI achieves a favorable trade-off between generative perplexity and entropy when varying the temperature.
Our sampling does not vary the temperature, relying solely on nucleus sampling with top-$p$=0.9, following the setup of~\citep{wang_remasking_2025}. 
With this setup, our SM yields a superior perplexity-entropy trade-off, as shown in Figure~\ref{fig:candi_comp}. 

\citet{jo_loopholing_2026} describe the information loss in binary MDLMs as a sampling wall, and propose to break this wall with their Loopholing method. 
Loopholing directly feeds the denoising model's output latent representation (prior to readout with embedding table). 
The denoising model is then conditioned on the superposition of the current tokens (masked and decoded ones). 
Similar to SM, their training relies on self-conditioning with a conditioning probability $p$. 
However, Loopholing has not been successful in pretraining continuation, requiring training from scratch. 
We hypothesize this limitation stems from feeding back the final latents directly (causing an embedding space mismatch) without any scaling.
In contrast, our learnable feedback operates in a known embedding space, enabling pretraining continuation and fine-tuning.
Consequently, Loopholing results are limited to small-scale language modeling and reasoning tasks.
On OWT, they achieve a notable test perplexity improvement (21.9), which is nonetheless slightly worse than our SM (21.47) (see Table~\ref{tab:owt-perplexity}).
Their results on unconditional text generation are not directly comparable to ours (showing overall much higher generative perplexity), which is likely due to their removal of time conditioning.

\citet{zhou_coevolutionary_2025} theoretically show that continuous diffusion models are more expressive than masked ones, arguing that the continuous model's achievable state trajectories are a superset of those of masked model.
In order to improve MDLMs, which are more popular in practice, Coevolutionary Continuous Discrete Diffusion (CCDD) proposes mixing discrete and continuous diffusion via a product of the two forward distributions.
The denoising model takes as input the concatenation of the discrete and continuous representations (along the sequence dimension), predicting both the continuous and discrete representations.
This doubles the input sequence length, thereby increasing compute and memory demands during training and inference.
In contrast, our SM merges continuous and discrete representations via superposition with minimal computational overhead (see Appendix~\ref{app:speed}).

Similarly, Latent Discrete Diffusion Models (LDDMs)~\citep{shariatian_latent_2025} enrich MDLMs with latents. 
The discrete and latent representation can either evolve simultaneously (FUlly JoInt Denoising; FUJI) or sequentially with the latent first (SEQuential denoising; SEQ). 
The two channels (discrete and latent) are interconnected through a multi-modal model or additional MLPs, which increase the computational complexity. 

Finally, dInfer~\citep{ma_dinfer_2025} introduces iteration smoothing, which adds a convex combination of softmax probability to the masked tokens. 
The weighting of the feedback is increased over the decoding iterations based on a fixed schedule. 
While elaborate ablations are missing, authors report an effective decrease of 30-40\% of decoding iterations thanks to iteration smoothing. 

\end{document}

%% file: math_commands.tex
\usepackage{amsmath,amsfonts,bm}

\def\Figref#1{Figure~\ref{#1}}

\def\eqref#1{equation~\ref{#1}}

\def\1{\bm{1}}

\DeclareMathAlphabet{\mathsfit}{\encodingdefault}{\sfdefault}{m}{sl}
\SetMathAlphabet{\mathsfit}{bold}{\encodingdefault}{\sfdefault}{bx}{n}

\newcommand{\E}{\mathbb{E}}
\newcommand{\Ls}{\mathcal{L}}



%% file: tables/k_experiment.tex
\begin{table}[h!]
\centering
\begin{tabular}{ccccccccc}
\toprule
     &                                                      & \multicolumn{2}{c}{Binary feedback} & \multicolumn{5}{c}{{SM feedback with top-$k$}} \\\cmidrule(r){3-4}\cmidrule(r){5-9}
               &     & No FT & {33.5 FT steps} & \multicolumn{5}{c}{{33.5k FT steps}}      \\\cmidrule(r){3-3}\cmidrule(r){4-4}\cmidrule(r){5-9}
Task & \begin{tabular}[c]{@{}c@{}}NFE\\ budget\end{tabular} & -                & -                & $k=1$  & $k=3$  & $k=5$  & $k=10$  & $k=|\mathcal{V}|$ \\\cmidrule(r){1-2}\cmidrule(r){3-4}\cmidrule(r){5-9}
          & 1/4 & 18.9  & 17.7 & 25.6 & 23.8 & 24.4 & 20.1 & 25.0 \\
Humaneval & 1/2 & 31.1  & 33.5 & 35.4 & 36.0 & 35.4 & 34.1 & 34.1 \\
          & 1/1 & 57.9  & 57.3 & 60.4 & 56.1 & 60.4 & 63.4 & 58.5 \\\cmidrule(r){1-2}\cmidrule(r){3-4}\cmidrule(r){5-9}
          & 1/4 & 26.6  & 27.2 & 32.8 & 31.4 & 27.8 & 27.6 & 27.2 \\
MBPP      & 1/2 & 42.8  & 43.6 & 49.6 & 48.0 & 45.4 & 45.6 & 44.4 \\
          & 1/1 & 57.8  & 58.4 & 56.2 & 57.6 & 57.2 & 56.8 & 57.8 \\\cmidrule(r){1-2}\cmidrule(r){3-4}\cmidrule(r){5-9}
          & 1/4 & 57.8  & 60.0 & 63.3 & 62.9 & 63.3 & 60.9 & 62.3 \\
GSM8k     & 1/2 & 76.0  & 78.7 & 80.3 & 79.8 & 80.9 & 81.7 & 78.9 \\
          & 1/1 & 82.0  & 83.5 & 82.9 & 84.6 & 84.2 & 83.6 & 84.0 \\\cmidrule(r){1-2}\cmidrule(r){3-4}\cmidrule(r){5-9}
          & 1/4 & 14.2  & 18.6 & 21.8 & 22.0 & 20.6 & 19.2 & 20.2 \\
Math-500  & 1/2 & 36.6  & 37.0 & 35.4 & 41.4 & 40.4 & 39.0 & 37.2 \\
          & 1/1 & 45.6  & 42.0 & 38.4 & 41.2 & 43.2 & 43.2 & 44.4 \\\cmidrule(r){1-2}\cmidrule(r){3-4}\cmidrule(r){5-9}
Avg.      & All & 45.6  & 46.5 & 48.5 & \textbf{48.7} & \underline{48.6} & 47.9 & 47.8\\
\bottomrule
\end{tabular}
\caption{
Comparison of different finetuned {(FT)} models, each trained with a different $k$ value. The $k=|\mathcal{V}|$ ablation is trained and evaluated with a learnable softmax temperature. Evaluations are performed on all three computational budgets and four math and coding evaluation tasks. The binary baselines are also included for comparison. We see that $k=1, 3$ and $5$ perform best, with degrading performance at higher $k$ values. The best performing model is highlighted in bold, and the second best is underlined.
}
\label{tab:k_experiment}
\end{table}

%% file: tables/time-dependence.tex
\begin{table}[h!]
\centering
\begin{tabular}{ccccc}
\toprule
     &                                              \multicolumn{2}{c}{{Binary feedback}} & \multicolumn{2}{c}{{SM feedback TD}} \\
     \cmidrule(r){2-3}\cmidrule(r){4-5}
               &      {No FT} & {33.5 FT steps} & \multicolumn{2}{c}{{33.5k FT steps}}      \\
               \cmidrule(r){2-2}\cmidrule(r){3-3}\cmidrule(r){4-5}
Task & -                & -                & Binary$\rightarrow$SM  & SM$\rightarrow$Binary \\\cmidrule(r){1-1}\cmidrule(r){2-3}\cmidrule(r){4-5}
Humaneval & $18.9$& $19.0_{(\pm 1.7)}$& $17.4_{(\pm 0.9)}$& $\mathbf{24.8}_{(\pm 1.1)}$\\
MBPP      & $26.6$& $27.0_{(\pm 1.6)}$& $25.2_{(\pm 1.3)}$& $\mathbf{30.9}_{(\pm 0.5)}$\\
GSM8k     & $57.8$& $59.5_{(\pm 1.1)}$& $59.3_{(\pm 1.1)}$& $\mathbf{61.6}_{(\pm 2.5)}$\\
Math-500  & $14.2$& $17.3_{(\pm 1.2)}$& $18.2_{(\pm 0.8)}$& $\mathbf{20.2}_{(\pm 1.9)}$\\
 \midrule
Avg.      & $29.4$& $31.4$&	$30.0$& $\mathbf{34.4}$\\ \bottomrule
\end{tabular}
\caption{Comparison of whether SM is more beneficial at the early or later stages of the denoising process. Both ablations Binary$\rightarrow$SM and SM$\rightarrow$Binary involve a denoising process in which 80\% of the steps are with SM and the remaining 20\% are with binary masking. We find that placing the SM steps at the beginning of denoising results in a much greater performance boost. These ablations are compared with our fully binary baselines. The two ablation columns are the \textit{mean} performance of the finetuned SM models with $k$=3 and a stepwise time-dependence. For each task, the best performing model is highlighted in bold. These evaluations are performed at an NFE budget of 1/4.}
\label{tab:time_dependence}
\end{table}

%% file: tables/time-dependence-early.tex
\begin{table}[h!]
\centering
\setlength{\tabcolsep}{4pt}
\begin{tabular}{cccccccc}
\toprule
& \multicolumn{2}{c}{{Binary feedback}} & \multicolumn{5}{c}{{SM feedback with top-$k$}} \\\cmidrule(r){2-3}\cmidrule(r){4-8}
                  & {No FT} & {33.5 FT steps} & \multicolumn{5}{c}{{33.5k FT steps}}      \\
               \cmidrule(r){2-2}\cmidrule(r){3-3}\cmidrule(r){4-8}             
Task & No TD & No TD & No TD & Linear  & $t' = 0.2$ & $t' = 0.5$ & $t' = 0.8$ \\\cmidrule(r){1-1}\cmidrule(r){2-3}\cmidrule(r){4-8}
Humaneval & $18.9$& $19.0_{(\pm 1.7)}$& $\mathbf{24.8}_{(\pm 1.1)}$& $22.4_{(\pm 1.2)}$& $\mathbf{24.8}_{(\pm 1.1)}$& $\mathbf{24.8}_{(\pm 1.1)}$& $\mathbf{24.8}_{(\pm 1.1)}$\\
MBPP     &  $26.6$& $27.0_{(\pm 1.6)}$& $30.8_{(\pm 0.5)}$& $28.4_{(\pm 1.0)}$& $\mathbf{30.9}_{(\pm 0.5)}$& $30.7_{(\pm 0.5)}$& $30.6_{(\pm 0.6)}$\\ 
GSM8k    & $57.8$& $59.5_{(\pm 1.1)}$& $\mathbf{62.3}_{(\pm 2.3)}$& $61.6_{(\pm 0.8)}$& $61.6_{(\pm 2.5)}$& $61.8_{(\pm 2.2)}$& $59.8_{(\pm 1.9)}$\\
Math-500  & $14.2$& $17.3_{(\pm 1.2)}$& $19.8_{(\pm 2.1)}$& $\mathbf{20.4}_{(\pm 1.4)}$& $20.2_{(\pm 1.9)}$& $19.8_{(\pm 1.4)}$& $18.2_{(\pm 1.8)}$\\
\midrule
Avg.      & $29.4$& $31.4$&	$\mathbf{34.4}$&	$33.2$&	$\mathbf{34.4}$&	$34.3$& $33.3$\\ \bottomrule
\end{tabular}
\caption{Comparison of varying forms of TD. All TD functions transition from SM$\rightarrow$Binary with different processes. These ablations are compared with our binary baselines. All SM ablation columns are the \textit{mean} performance of the finetuned SM models with $k$=3 and the associated TD. For each task, the best performing model is highlighted in bold.}
\label{tab:time_dependence_early}
\end{table}

%% file: main.bbl
\begin{thebibliography}{74}
\providecommand{\natexlab}[1]{#1}
\providecommand{\url}[1]{\texttt{#1}}
\expandafter\ifx\csname urlstyle\endcsname\relax
  \providecommand{\doi}[1]{doi: #1}\else
  \providecommand{\doi}{doi: \begingroup \urlstyle{rm}\Url}\fi

\bibitem[Allal et~al.(2025)Allal, Lozhkov, Bakouch, Blázquez, Penedo, Tunstall, Marafioti, Kydlíček, Lajarín, Srivastav, Lochner, Fahlgren, Nguyen, Fourrier, Burtenshaw, Larcher, Zhao, Zakka, Morlon, Raffel, von Werra, and Wolf]{smol_2025}
Loubna~Ben Allal, Anton Lozhkov, Elie Bakouch, Gabriel~Martín Blázquez, Guilherme Penedo, Lewis Tunstall, Andrés Marafioti, Hynek Kydlíček, Agustín~Piqueres Lajarín, Vaibhav Srivastav, Joshua Lochner, Caleb Fahlgren, Xuan-Son Nguyen, Clémentine Fourrier, Ben Burtenshaw, Hugo Larcher, Haojun Zhao, Cyril Zakka, Mathieu Morlon, Colin Raffel, Leandro von Werra, and Thomas Wolf.
\newblock {SmolLM2}: When smol goes big -- data-centric training of a small language model.
\newblock \emph{arXiv preprint arXiv:2502.02737}, 2025.
\newblock URL \url{https://arxiv.org/abs/2502.02737}.

\bibitem[Arriola et~al.(2024)Arriola, Gokaslan, Chiu, Yang, Qi, Han, Sahoo, and Kuleshov]{arriola_block_2024}
Marianne Arriola, Aaron Gokaslan, Justin~T. Chiu, Zhihan Yang, Zhixuan Qi, Jiaqi Han, Subham~Sekhar Sahoo, and Volodymyr Kuleshov.
\newblock Block {Diffusion}: {Interpolating} {Between} {Autoregressive} and {Diffusion} {Language} {Models}.
\newblock In \emph{The {Thirteenth} {International} {Conference} on {Learning} {Representations} ({ICLR})}, October 2024.
\newblock URL \url{https://openreview.net/forum?id=tyEyYT267x}.

\bibitem[Austin et~al.(2021{\natexlab{a}})Austin, Johnson, Ho, Tarlow, and Berg]{austin_structured_2021}
Jacob Austin, Daniel~D. Johnson, Jonathan Ho, Daniel Tarlow, and Rianne van~den Berg.
\newblock Structured {Denoising} {Diffusion} {Models} in {Discrete} {State}-{Spaces}.
\newblock In \emph{Advances in {Neural} {Information} {Processing} {Systems} ({NeurIPS})}, volume~34, November 2021{\natexlab{a}}.
\newblock URL \url{https://openreview.net/forum?id=h7-XixPCAL}.

\bibitem[Austin et~al.(2021{\natexlab{b}})Austin, Odena, Nye, Bosma, Michalewski, Dohan, Jiang, Cai, Terry, Le, and Sutton]{mbpp_2021}
Jacob Austin, Augustus Odena, Maxwell Nye, Maarten Bosma, Henryk Michalewski, David Dohan, Ellen Jiang, Carrie Cai, Michael Terry, Quoc Le, and Charles Sutton.
\newblock Program synthesis with large language models.
\newblock \emph{arXiv preprint arXiv:2108.07732}, 2021{\natexlab{b}}.
\newblock URL \url{https://arxiv.org/abs/2108.07732}.

\bibitem[Black et~al.(2023)Black, Janner, Du, Kostrikov, and Levine]{black_training_2023}
Kevin Black, Michael Janner, Yilun Du, Ilya Kostrikov, and Sergey Levine.
\newblock Training {Diffusion} {Models} with {Reinforcement} {Learning}.
\newblock In \emph{The {Twelfth} {International} {Conference} on {Learning} {Representations} ({ICLR})}, October 2023.
\newblock URL \url{https://openreview.net/forum?id=YCWjhGrJFD}.

\bibitem[Brown et~al.(2020)Brown, Mann, Ryder, Subbiah, Kaplan, Dhariwal, Neelakantan, Shyam, Sastry, Askell, Agarwal, Herbert-Voss, Krueger, Henighan, Child, Ramesh, Ziegler, Wu, Winter, Hesse, Chen, Sigler, Litwin, Gray, Chess, Clark, Berner, McCandlish, Radford, Sutskever, and Amodei]{brown_language_2020}
Tom Brown, Benjamin Mann, Nick Ryder, Melanie Subbiah, Jared~D Kaplan, Prafulla Dhariwal, Arvind Neelakantan, Pranav Shyam, Girish Sastry, Amanda Askell, Sandhini Agarwal, Ariel Herbert-Voss, Gretchen Krueger, Tom Henighan, Rewon Child, Aditya Ramesh, Daniel Ziegler, Jeffrey Wu, Clemens Winter, Chris Hesse, Mark Chen, Eric Sigler, Mateusz Litwin, Scott Gray, Benjamin Chess, Jack Clark, Christopher Berner, Sam McCandlish, Alec Radford, Ilya Sutskever, and Dario Amodei.
\newblock Language {Models} are {Few}-{Shot} {Learners}.
\newblock In \emph{Advances in {Neural} {Information} {Processing} {Systems} ({NeurIPS})}, volume~33, 2020.
\newblock URL \url{https://proceedings.neurips.cc/paper_files/paper/2020/file/1457c0d6bfcb4967418bfb8ac142f64a-Paper.pdf}.

\bibitem[Campbell et~al.(2022)Campbell, Benton, and Bortoli]{campbell_continuous_2022}
Andrew Campbell, Joe Benton, and Valentin~De Bortoli.
\newblock A {Continuous} {Time} {Framework} for {Discrete} {Denoising} {Models}.
\newblock In \emph{Advances in {Neural} {Information} {Processing} {Systems} ({NeurIPS})}, volume~35, 2022.
\newblock URL \url{https://openreview.net/pdf?id=DmT862YAieY}.

\bibitem[Chao et~al.(2025)Chao, Sun, Liang, Lee, and Krishnan]{chao_masked_2025}
Chen-Hao Chao, Wei-Fang Sun, Hanwen Liang, Chun-Yi Lee, and Rahul~G. Krishnan.
\newblock Beyond {Masked} and {Unmasked}: {Discrete} {Diffusion} {Models} via {Partial} {Masking}.
\newblock In \emph{The {Thirty}-ninth {Annual} {Conference} on {Neural} {Information} {Processing} {Systems} ({NeurIPS})}, 2025.
\newblock \doi{10.48550/arXiv.2505.18495}.
\newblock URL \url{https://openreview.net/forum?id=nqpbbvEZwF}.

\bibitem[Chen et~al.(2021)Chen, Tworek, Jun, Yuan, Pinto, Kaplan, Edwards, Burda, Joseph, Brockman, Ray, Puri, Krueger, Petrov, Khlaaf, Sastry, Mishkin, Chan, Gray, Ryder, Pavlov, Power, Kaiser, Bavarian, Winter, Tillet, Such, Cummings, Plappert, Chantzis, Barnes, Herbert-Voss, Guss, Nichol, Paino, Tezak, Tang, Babuschkin, Balaji, Jain, Saunders, Hesse, Carr, Leike, Achiam, Misra, Morikawa, Radford, Knight, Brundage, Murati, Mayer, Welinder, McGrew, Amodei, McCandlish, Sutskever, and Zaremba]{chen_evaluating_2021}
Mark Chen, Jerry Tworek, Heewoo Jun, Qiming Yuan, Henrique Ponde de~Oliveira Pinto, Jared Kaplan, Harri Edwards, Yuri Burda, Nicholas Joseph, Greg Brockman, Alex Ray, Raul Puri, Gretchen Krueger, Michael Petrov, Heidy Khlaaf, Girish Sastry, Pamela Mishkin, Brooke Chan, Scott Gray, Nick Ryder, Mikhail Pavlov, Alethea Power, Lukasz Kaiser, Mohammad Bavarian, Clemens Winter, Philippe Tillet, Felipe~Petroski Such, Dave Cummings, Matthias Plappert, Fotios Chantzis, Elizabeth Barnes, Ariel Herbert-Voss, William~Hebgen Guss, Alex Nichol, Alex Paino, Nikolas Tezak, Jie Tang, Igor Babuschkin, Suchir Balaji, Shantanu Jain, William Saunders, Christopher Hesse, Andrew~N. Carr, Jan Leike, Josh Achiam, Vedant Misra, Evan Morikawa, Alec Radford, Matthew Knight, Miles Brundage, Mira Murati, Katie Mayer, Peter Welinder, Bob McGrew, Dario Amodei, Sam McCandlish, Ilya Sutskever, and Wojciech Zaremba.
\newblock Evaluating {Large} {Language} {Models} {Trained} on {Code}.
\newblock \emph{arXiv preprint arXiv:2107.03374}, July 2021.
\newblock \doi{10.48550/arXiv.2107.03374}.
\newblock URL \url{http://arxiv.org/abs/2107.03374}.

\bibitem[Chen et~al.(2023)Chen, Zhang, and Hinton]{chen_analog_2023}
Ting Chen, Ruixiang Zhang, and Geoffrey Hinton.
\newblock Analog {Bits}: {Generating} {Discrete} {Data} using {Diffusion} {Models} with {Self}-{Conditioning}.
\newblock In \emph{The {Eleventh} {International} {Conference} on {Learning} {Representations} ({ICLR})}, 2023.
\newblock URL \url{https://openreview.net/forum?id=3itjR9QxFw}.

\bibitem[Cobbe et~al.(2021)Cobbe, Kosaraju, Bavarian, Chen, Jun, Kaiser, Plappert, Tworek, Hilton, Nakano, Hesse, and Schulman]{cobbe_training_2021}
Karl Cobbe, Vineet Kosaraju, Mohammad Bavarian, Mark Chen, Heewoo Jun, Lukasz Kaiser, Matthias Plappert, Jerry Tworek, Jacob Hilton, Reiichiro Nakano, Christopher Hesse, and John Schulman.
\newblock Training {Verifiers} to {Solve} {Math} {Word} {Problems}.
\newblock \emph{arXiv preprint arXiv:2110.14168}, November 2021.
\newblock \doi{10.48550/arXiv.2110.14168}.
\newblock URL \url{http://arxiv.org/abs/2110.14168}.

\bibitem[Codefuse \& Team(2025)Codefuse and Team]{codefuse2025samplemattersleveragingmixtureofexperts}
Codefuse and Ling Team.
\newblock Every sample matters: Leveraging mixture-of-experts and high-quality data for efficient and accurate code {LLM}.
\newblock \emph{arXiv preprint arXiv:2503.17793}, 2025.
\newblock URL \url{https://arxiv.org/abs/2503.17793}.

\bibitem[DeepMind(2025)]{deepmind_gemini_2025}
DeepMind.
\newblock Gemini {Diffusion}, 2025.
\newblock URL \url{https://deepmind.google/models/gemini-diffusion/}.

\bibitem[Dieleman et~al.(2022)Dieleman, Sartran, Roshannai, Savinov, Ganin, Richemond, Doucet, Strudel, Dyer, Durkan, Hawthorne, Leblond, Grathwohl, and Adler]{dieleman_continuous_2022}
Sander Dieleman, Laurent Sartran, Arman Roshannai, Nikolay Savinov, Yaroslav Ganin, Pierre~H. Richemond, Arnaud Doucet, Robin Strudel, Chris Dyer, Conor Durkan, Curtis Hawthorne, Rémi Leblond, Will Grathwohl, and Jonas Adler.
\newblock Continuous diffusion for categorical data.
\newblock \emph{arXiv preprint arXiv:2211.15089}, December 2022.
\newblock \doi{10.48550/arXiv.2211.15089}.
\newblock URL \url{http://arxiv.org/abs/2211.15089}.

\bibitem[Geiping et~al.(2025)Geiping, McLeish, Jain, Kirchenbauer, Singh, Bartoldson, Kailkhura, Bhatele, and Goldstein]{geiping_scaling_2025}
Jonas Geiping, Sean McLeish, Neel Jain, John Kirchenbauer, Siddharth Singh, Brian~R. Bartoldson, Bhavya Kailkhura, Abhinav Bhatele, and Tom Goldstein.
\newblock Scaling up {Test}-{Time} {Compute} with {Latent} {Reasoning}: {A} {Recurrent} {Depth} {Approach}.
\newblock In \emph{{ES}-{FoMo} {III}: 3rd {Workshop} on {Efficient} {Systems} for {Foundation} {Models}}, 2025.
\newblock URL \url{https://openreview.net/forum?id=D6o6Bwtq7h}.

\bibitem[Gokaslan \& Cohen(2019)Gokaslan and Cohen]{Gokaslan2019OpenWeb}
Aaron Gokaslan and Vanya Cohen.
\newblock Openwebtext corpus.
\newblock \url{http://Skylion007.github.io/OpenWebTextCorpus}, 2019.

\bibitem[Gong et~al.(2022)Gong, Li, Feng, Wu, and Kong]{gong_diffuseq_2022}
Shansan Gong, Mukai Li, Jiangtao Feng, Zhiyong Wu, and Lingpeng Kong.
\newblock {DiffuSeq}: {Sequence} to {Sequence} {Text} {Generation} with {Diffusion} {Models}.
\newblock In \emph{The {Eleventh} {International} {Conference} on {Learning} {Representations} ({ICLR})}, September 2022.
\newblock URL \url{https://openreview.net/forum?id=jQj-_rLVXsj}.

\bibitem[Gong et~al.(2023)Gong, Li, Feng, Wu, and Kong]{gong_diffuseqv2_2023}
Shansan Gong, Mukai Li, Jiangtao Feng, Zhiyong Wu, and Lingpeng Kong.
\newblock {DiffuSeq}-v2: {Bridging} {Discrete} and {Continuous} {Text} {Spaces} for {Accelerated} {Seq2Seq} {Diffusion} {Models}.
\newblock In \emph{Findings of the {Association} for {Computational} {Linguistics}: {EMNLP} 2023}, December 2023.
\newblock \doi{10.18653/v1/2023.findings-emnlp.660}.
\newblock URL \url{https://aclanthology.org/2023.findings-emnlp.660/}.

\bibitem[Gong et~al.(2025{\natexlab{a}})Gong, Agarwal, Zhang, Ye, Zheng, Li, An, Zhao, Bi, Han, Peng, and Kong]{gong_scaling_2025}
Shansan Gong, Shivam Agarwal, Yizhe Zhang, Jiacheng Ye, Lin Zheng, Mukai Li, Chenxin An, Peilin Zhao, Wei Bi, Jiawei Han, Hao Peng, and Lingpeng Kong.
\newblock Scaling {Diffusion} {Language} {Models} via {Adaptation} from {Autoregressive} {Models}.
\newblock In \emph{The {Thirteenth} {International} {Conference} on {Learning} {Representations} ({ICLR})}, May 2025{\natexlab{a}}.
\newblock URL \url{https://openreview.net/forum?id=j1tSLYKwg8}.

\bibitem[Gong et~al.(2025{\natexlab{b}})Gong, Zhang, Zheng, Gu, Jaitly, Kong, and Zhang]{gong_diffucoder_2025}
Shansan Gong, Ruixiang Zhang, Huangjie Zheng, Jiatao Gu, Navdeep Jaitly, Lingpeng Kong, and Yizhe Zhang.
\newblock {DiffuCoder}: {Understanding} and {Improving} {Masked} {Diffusion} {Models} for {Code} {Generation}.
\newblock \emph{arXiv preprint arXiv:2506.20639}, June 2025{\natexlab{b}}.
\newblock \doi{10.48550/arXiv.2506.20639}.
\newblock URL \url{http://arxiv.org/abs/2506.20639}.

\bibitem[Gulrajani \& Hashimoto(2023)Gulrajani and Hashimoto]{gulrajani_likelihoodbased_2023}
Ishaan Gulrajani and Tatsunori Hashimoto.
\newblock Likelihood-{Based} {Diffusion} {Language} {Models}.
\newblock In \emph{Advances in {Neural} {Information} {Processing} {Systems} ({NeurIPS})}, volume~37, November 2023.
\newblock URL \url{https://openreview.net/forum?id=e2MCL6hObn&noteId=ueUWS1aqtE}.

\bibitem[Guo et~al.(2025)Guo, Yang, Zhang, Song, Wang, Zhu, Xu, Zhang, Ma, Bi, Zhang, Yu, Wu, Wu, Gou, Shao, Li, Gao, et~al.]{deepseekai2025short}
Daya Guo, Dejian Yang, Haowei Zhang, Junxiao Song, Peiyi Wang, Qihao Zhu, Runxin Xu, Ruoyu Zhang, Shirong Ma, Xiao Bi, Xiaokang Zhang, Xingkai Yu, Yu~Wu, Z.~F. Wu, Zhibin Gou, Zhihong Shao, Zhuoshu Li, Ziyi Gao, et~al.
\newblock {DeepSeek}-{R1} incentivizes reasoning in {LLMs} through reinforcement learning.
\newblock \emph{Nature}, 645\penalty0 (8081):\penalty0 633--638, September 2025.
\newblock ISSN 1476-4687.
\newblock \doi{10.1038/s41586-025-09422-z}.
\newblock URL \url{https://doi.org/10.1038/s41586-025-09422-z}.

\bibitem[Hao et~al.(2025)Hao, Sukhbaatar, Su, Li, Hu, Weston, and Tian]{hao_training_2025}
Shibo Hao, Sainbayar Sukhbaatar, DiJia Su, Xian Li, Zhiting Hu, Jason Weston, and Yuandong Tian.
\newblock Training {Large} {Language} {Models} to {Reason} in a {Continuous} {Latent} {Space}.
\newblock In \emph{Second {Conference} on {Language} {Modeling}}, 2025.
\newblock \doi{10.48550/arXiv.2412.06769}.
\newblock URL \url{https://openreview.net/forum?id=Itxz7S4Ip3}.

\bibitem[Ho et~al.(2020)Ho, Jain, and Abbeel]{ho_denoising_2020}
Jonathan Ho, Ajay Jain, and Pieter Abbeel.
\newblock Denoising {Diffusion} {Probabilistic} {Models}.
\newblock In \emph{Advances in {Neural} {Information} {Processing} {Systems} ({NeurIPS})}, volume~34, 2020.
\newblock URL \url{https://proceedings.neurips.cc/paper/2020/file/4c5bcfec8584af0d967f1ab10179ca4b-Paper.pdf}.

\bibitem[Hui et~al.(2024)Hui, Yang, Cui, Yang, Liu, Zhang, Liu, Zhang, Yu, Lu, Dang, Fan, Zhang, Yang, Men, Huang, Zheng, Miao, Quan, Feng, Ren, Ren, Zhou, and Lin]{hui_qwen25coder_2024}
Binyuan Hui, Jian Yang, Zeyu Cui, Jiaxi Yang, Dayiheng Liu, Lei Zhang, Tianyu Liu, Jiajun Zhang, Bowen Yu, Keming Lu, Kai Dang, Yang Fan, Yichang Zhang, An~Yang, Rui Men, Fei Huang, Bo~Zheng, Yibo Miao, Shanghaoran Quan, Yunlong Feng, Xingzhang Ren, Xuancheng Ren, Jingren Zhou, and Junyang Lin.
\newblock Qwen2.5-{Coder} {Technical} {Report}.
\newblock \emph{arXiv preprint arXiv:2409.12186}, November 2024.
\newblock \doi{10.48550/arXiv.2409.12186}.
\newblock URL \url{http://arxiv.org/abs/2409.12186}.

\bibitem[Inception et~al.(2025)Inception, Khanna, Kharbanda, Li, Varma, Wang, Birnbaum, Luo, Miraoui, Palrecha, Ermon, Grover, and Kuleshov]{inception_mercury_2025}
Labs Inception, Samar Khanna, Siddhant Kharbanda, Shufan Li, Harshit Varma, Eric Wang, Sawyer Birnbaum, Ziyang Luo, Yanis Miraoui, Akash Palrecha, Stefano Ermon, Aditya Grover, and Volodymyr Kuleshov.
\newblock Mercury: {Ultra}-{Fast} {Language} {Models} {Based} on {Diffusion}.
\newblock \emph{arXiv preprint arXiv:2506.17298}, 2025.
\newblock \doi{10.48550/arXiv.2506.17298}.
\newblock URL \url{http://arxiv.org/abs/2506.17298}.

\bibitem[Jin et~al.(2025)Jin, Wang, Gao, Wen, Qi, Liu, and Zhang]{jin_thinking_2025}
Xiangqi Jin, Yuxuan Wang, Yifeng Gao, Zichen Wen, Biqing Qi, Dongrui Liu, and Linfeng Zhang.
\newblock Thinking {Inside} the {Mask}: {In}-{Place} {Prompting} in {Diffusion} {LLMs}.
\newblock \emph{arXiv preprint arXiv:2508.10736}, August 2025.
\newblock \doi{10.48550/arXiv.2508.10736}.
\newblock URL \url{http://arxiv.org/abs/2508.10736}.

\bibitem[Jo et~al.(2026)Jo, Yoon, Deschenaux, Gulcehre, and Ahn]{jo_loopholing_2026}
Mingyu Jo, Jaesik Yoon, Justin Deschenaux, Caglar Gulcehre, and Sungjin Ahn.
\newblock Loopholing {Discrete} {Diffusion}: {Deterministic} {Bypass} of the {Sampling} {Wall}.
\newblock In \emph{The {Fourteenth} {International} {Conference} on {Learning} {Representations} ({ICLR})}, 2026.
\newblock \doi{10.48550/arXiv.2510.19304}.
\newblock URL \url{https://openreview.net/forum?id=wM3Mdk2Pa5}.

\bibitem[Lambert et~al.(2025)Lambert, Morrison, Pyatkin, Huang, Ivison, Brahman, Miranda, Liu, Dziri, Lyu, Gu, Malik, Graf, Hwang, Yang, Bras, Tafjord, Wilhelm, Soldaini, Smith, Wang, Dasigi, and Hajishirzi]{tulu3_2025}
Nathan Lambert, Jacob Morrison, Valentina Pyatkin, Shengyi Huang, Hamish Ivison, Faeze Brahman, Lester James~V. Miranda, Alisa Liu, Nouha Dziri, Shane Lyu, Yuling Gu, Saumya Malik, Victoria Graf, Jena~D. Hwang, Jiangjiang Yang, Ronan~Le Bras, Oyvind Tafjord, Chris Wilhelm, Luca Soldaini, Noah~A. Smith, Yizhong Wang, Pradeep Dasigi, and Hannaneh Hajishirzi.
\newblock Tulu 3: Pushing frontiers in open language model post-training.
\newblock \emph{arXiv preprint arXiv:2411.15124}, 2025.
\newblock URL \url{https://arxiv.org/abs/2411.15124}.

\bibitem[{LCM Team} et~al.(2024){LCM Team}, Barrault, Duquenne, Elbayad, Kozhevnikov, Alastruey, Andrews, Coria, Couairon, Costa-jussà, Dale, Elsahar, Heffernan, Janeiro, Tran, Ropers, Sánchez, Roman, Mourachko, Saleem, and Schwenk]{lcm_large_2024}
{LCM Team}, Loïc Barrault, Paul-Ambroise Duquenne, Maha Elbayad, Artyom Kozhevnikov, Belen Alastruey, Pierre Andrews, Mariano Coria, Guillaume Couairon, Marta~R. Costa-jussà, David Dale, Hady Elsahar, Kevin Heffernan, João~Maria Janeiro, Tuan Tran, Christophe Ropers, Eduardo Sánchez, Robin~San Roman, Alexandre Mourachko, Safiyyah Saleem, and Holger Schwenk.
\newblock Large {Concept} {Models}: {Language} {Modeling} in a {Sentence} {Representation} {Space}.
\newblock \emph{arXiv preprint arXiv:2412.08821}, December 2024.
\newblock \doi{10.48550/arXiv.2412.08821}.
\newblock URL \url{http://arxiv.org/abs/2412.08821}.

\bibitem[Li et~al.(2022)Li, Thickstun, Gulrajani, Liang, and Hashimoto]{li_diffusionlm_2022}
Xiang~Lisa Li, John Thickstun, Ishaan Gulrajani, Percy Liang, and Tatsunori Hashimoto.
\newblock Diffusion-{LM} {Improves} {Controllable} {Text} {Generation}.
\newblock In \emph{Advances in {Neural} {Information} {Processing} {Systems} ({NeurIPS})}, volume~36, October 2022.
\newblock URL \url{https://openreview.net/forum?id=3s9IrEsjLyk}.

\bibitem[Lightman et~al.(2024)Lightman, Kosaraju, Burda, Edwards, Baker, Lee, Leike, Schulman, Sutskever, and Cobbe]{lightman2024math}
Hunter Lightman, Vineet Kosaraju, Yuri Burda, Harrison Edwards, Bowen Baker, Teddy Lee, Jan Leike, John Schulman, Ilya Sutskever, and Karl Cobbe.
\newblock Let's verify step by step.
\newblock In \emph{The Twelfth International Conference on Learning Representations (ICLR)}, 2024.
\newblock URL \url{https://openreview.net/forum?id=v8L0pN6EOi}.

\bibitem[Liu et~al.(2023)Liu, Xia, Wang, and Zhang]{evalplus_2023}
Jiawei Liu, Chunqiu~Steven Xia, Yuyao Wang, and Lingming Zhang.
\newblock Is your code generated by chat{GPT} really correct? rigorous evaluation of large language models for code generation.
\newblock In \emph{Thirty-seventh Conference on Neural Information Processing Systems (NeurIPS)}, 2023.
\newblock URL \url{https://openreview.net/forum?id=1qvx610Cu7}.

\bibitem[Liu et~al.(2024)Liu, Wang, Yin, Molchanov, Wang, Cheng, and Chen]{shih_dora_2024}
Shih-Yang Liu, Chien-Yi Wang, Hongxu Yin, Pavlo Molchanov, Yu-Chiang~Frank Wang, Kwang-Ting Cheng, and Min-Hung Chen.
\newblock {DoRA}: Weight-decomposed low-rank adaptation.
\newblock \emph{arXiv preprint arXiv:2402.09353}, 2024.
\newblock URL \url{https://arxiv.org/abs/2402.09353}.

\bibitem[Liu et~al.(2025)Liu, Yang, Zhang, Chen, Zou, Wei, Wang, and Zhang]{liu_dllmcache_2025}
Zhiyuan Liu, Yicun Yang, Yaojie Zhang, Junjie Chen, Chang Zou, Qingyuan Wei, Shaobo Wang, and Linfeng Zhang.
\newblock {dLLM}-{Cache}: {Accelerating} {Diffusion} {Large} {Language} {Models} with {Adaptive} {Caching}.
\newblock \emph{arXiv preprint arXiv:2506.06295}, May 2025.
\newblock \doi{10.48550/arXiv.2506.06295}.
\newblock URL \url{http://arxiv.org/abs/2506.06295}.

\bibitem[Lou et~al.(2024)Lou, Meng, and Ermon]{lou_discrete_2024}
Aaron Lou, Chenlin Meng, and Stefano Ermon.
\newblock Discrete diffusion modeling by estimating the ratios of the data distribution.
\newblock In \emph{Proceedings of the 41st {International} {Conference} on {Machine} {Learning} ({ICML})}, volume 235, July 2024.
\newblock URL \url{https://openreview.net/forum?id=CNicRIVIPA}.

\bibitem[Ma et~al.(2025)Ma, Du, Wei, Chen, Xu, Wang, Feng, Lu, Liu, Qi, Zhang, Tao, Feng, Jiang, Xu, Huang, Zhuang, Xu, Hu, Lan, Zhao, Li, and Zheng]{ma_dinfer_2025}
Yuxin Ma, Lun Du, Lanning Wei, Kun Chen, Qian Xu, Kangyu Wang, Guofeng Feng, Guoshan Lu, Lin Liu, Xiaojing Qi, Xinyuan Zhang, Zhen Tao, Haibo Feng, Ziyun Jiang, Ying Xu, Zenan Huang, Yihong Zhuang, Haokai Xu, Jiaqi Hu, Zhenzhong Lan, Junbo Zhao, Jianguo Li, and Da~Zheng.
\newblock {dInfer}: {An} {Efficient} {Inference} {Framework} for {Diffusion} {Language} {Models}.
\newblock \emph{arXiv preprint arXiv:2510.08666}, October 2025.
\newblock \doi{10.48550/arXiv.2510.08666}.
\newblock URL \url{http://arxiv.org/abs/2510.08666}.

\bibitem[Ni et~al.(2025)Ni, Liu, Dou, Du, Wang, Yan, Pang, and Shieh]{ni_diffusion_2025}
Jinjie Ni, Qian Liu, Longxu Dou, Chao Du, Zili Wang, Hang Yan, Tianyu Pang, and Michael~Qizhe Shieh.
\newblock Diffusion {Language} {Models} are {Super} {Data} {Learners}.
\newblock \emph{arXiv preprint arXiv:2511.03276}, November 2025.
\newblock \doi{10.48550/arXiv.2511.03276}.
\newblock URL \url{http://arxiv.org/abs/2511.03276}.

\bibitem[Nie et~al.(2025)Nie, Zhu, You, Zhang, Ou, Hu, Zhou, Lin, Wen, and Li]{nie_large_2025}
Shen Nie, Fengqi Zhu, Zebin You, Xiaolu Zhang, Jingyang Ou, Jun Hu, Jun Zhou, Yankai Lin, Ji-Rong Wen, and Chongxuan Li.
\newblock Large {Language} {Diffusion} {Models}.
\newblock In \emph{Advances in {Neural} {Information} {Processing} {Systems} ({NeurIPS})}, 2025.
\newblock URL \url{https://openreview.net/forum?id=KnqiC0znVF}.

\bibitem[OpenAI(2024)]{openai_learning_2024}
OpenAI.
\newblock Learning to reason with {LLMs}, September 2024.
\newblock URL \url{https://openai.com/index/learning-to-reason-with-llms/}.
\newblock https://openai.com/index/learning-to-reason-with-llms/.

\bibitem[Ou et~al.(2025)Ou, Nie, Xue, Zhu, Sun, Li, and Li]{ou_your_2025}
Jingyang Ou, Shen Nie, Kaiwen Xue, Fengqi Zhu, Jiacheng Sun, Zhenguo Li, and Chongxuan Li.
\newblock Your {Absorbing} {Discrete} {Diffusion} {Secretly} {Models} the {Conditional} {Distributions} of {Clean} {Data}.
\newblock In \emph{The {Thirteenth} {International} {Conference} on {Learning} {Representations} ({ICLR})}, 2025.
\newblock URL \url{https://openreview.net/forum?id=sMyXP8Tanm}.

\bibitem[Peebles \& Xie(2023)Peebles and Xie]{peebles_scalable_2023}
William Peebles and Saining Xie.
\newblock Scalable {Diffusion} {Models} with {Transformers}.
\newblock In \emph{Proceedings of the {IEEE}/{CVF} {International} {Conference} on {Computer} {Vision} ({CVPR})}, pp.\  4195--4205, 2023.
\newblock URL \url{https://openaccess.thecvf.com/content/ICCV2023/html/Peebles_Scalable_Diffusion_Models_with_Transformers_ICCV_2023_paper.html}.

\bibitem[Pillutla et~al.(2021)Pillutla, Swayamdipta, Zellers, Thickstun, Welleck, Choi, and Harchaoui]{pillutla2021mauve}
Krishna Pillutla, Swabha Swayamdipta, Rowan Zellers, John Thickstun, Sean Welleck, Yejin Choi, and Zaid Harchaoui.
\newblock {MAUVE}: Measuring the gap between neural text and human text using divergence frontiers.
\newblock In \emph{Advances in Neural Information Processing Systems (NeurIPS)}, 2021.
\newblock URL \url{https://openreview.net/forum?id=Tqx7nJp7PR}.

\bibitem[Prabhudesai et~al.(2025)Prabhudesai, Wu, Zadeh, Fragkiadaki, and Pathak]{prabhudesai_diffusion_2025}
Mihir Prabhudesai, Mengning Wu, Amir Zadeh, Katerina Fragkiadaki, and Deepak Pathak.
\newblock Diffusion {Beats} {Autoregressive} in {Data}-{Constrained} {Settings}.
\newblock In \emph{The {Thirty}-ninth {Annual} {Conference} on {Neural} {Information} {Processing} {Systems} ({NeurIPS})}, 2025.
\newblock URL \url{https://openreview.net/forum?id=W5Ht05jF4c}.

\bibitem[Pynadath et~al.(2025)Pynadath, Shi, and Zhang]{pynadath_candi_2025}
Patrick Pynadath, Jiaxin Shi, and Ruqi Zhang.
\newblock {CANDI}: {Hybrid} {Discrete}-{Continuous} {Diffusion} {Models}.
\newblock \emph{arXiv preprint arXiv:2510.22510}, October 2025.
\newblock \doi{10.48550/arXiv.2510.22510}.
\newblock URL \url{http://arxiv.org/abs/2510.22510}.

\bibitem[Qwen et~al.(2025)Qwen, Yang, Yang, Zhang, Hui, Zheng, Yu, Li, Liu, Huang, Wei, Lin, Yang, Tu, Zhang, Yang, Yang, Zhou, Lin, Dang, Lu, Bao, Yang, Yu, Li, Xue, Zhang, Zhu, Men, Lin, Li, Tang, Xia, Ren, Ren, Fan, Su, Zhang, Wan, Liu, Cui, Zhang, and Qiu]{qwen_qwen25_2025}
Qwen, An~Yang, Baosong Yang, Beichen Zhang, Binyuan Hui, Bo~Zheng, Bowen Yu, Chengyuan Li, Dayiheng Liu, Fei Huang, Haoran Wei, Huan Lin, Jian Yang, Jianhong Tu, Jianwei Zhang, Jianxin Yang, Jiaxi Yang, Jingren Zhou, Junyang Lin, Kai Dang, Keming Lu, Keqin Bao, Kexin Yang, Le~Yu, Mei Li, Mingfeng Xue, Pei Zhang, Qin Zhu, Rui Men, Runji Lin, Tianhao Li, Tianyi Tang, Tingyu Xia, Xingzhang Ren, Xuancheng Ren, Yang Fan, Yang Su, Yichang Zhang, Yu~Wan, Yuqiong Liu, Zeyu Cui, Zhenru Zhang, and Zihan Qiu.
\newblock Qwen2.5 {Technical} {Report}.
\newblock \emph{arXiv preprint arXiv:2412.15115}, January 2025.
\newblock \doi{10.48550/arXiv.2412.15115}.
\newblock URL \url{http://arxiv.org/abs/2412.15115}.

\bibitem[{Qwen Team}(2024)]{qwen_qwq_2024}
{Qwen Team}.
\newblock {QwQ}: {Reflect} {Deeply} on the {Boundaries} of the {Unknown}, November 2024.
\newblock URL \url{https://qwenlm.github.io/blog/qwq-32b-preview/}.
\newblock https://qwenlm.github.io/blog/qwq-32b-preview/.

\bibitem[Radford et~al.(2019)Radford, Wu, Child, Luan, Amodei, and Sutskever]{radford_2019_gpt2}
Alec Radford, Jeff Wu, Rewon Child, David Luan, Dario Amodei, and Ilya Sutskever.
\newblock Language models are unsupervised multitask learners, 2019.
\newblock URL \url{https://storage.prod.researchhub.com/uploads/papers/2020/06/01/language-models.pdf}.

\bibitem[Rütte et~al.(2025)Rütte, Fluri, Ding, Orvieto, Schölkopf, and Hofmann]{rutte_generalized_2025}
Dimitri~von Rütte, Janis Fluri, Yuhui Ding, Antonio Orvieto, Bernhard Schölkopf, and Thomas Hofmann.
\newblock Generalized {Interpolating} {Discrete} {Diffusion}.
\newblock In \emph{Forty-second {International} {Conference} on {Machine} {Learning} ({ICML})}, 2025.
\newblock URL \url{https://openreview.net/forum?id=rvZv7sDPV9}.

\bibitem[Sahoo et~al.(2024)Sahoo, Arriola, and Schiff]{sahoo_simple_2024}
Subham~Sekhar Sahoo, Marianne Arriola, and Yair Schiff.
\newblock Simple and {Effective} {Masked} {Diffusion} {Language} {Models}.
\newblock In \emph{Advances in {Neural} {Information} {Processing} {Systems} ({NeurIPS})}, volume~38, 2024.
\newblock URL \url{https://openreview.net/forum?id=L4uaAR4ArM}.

\bibitem[Sahoo et~al.(2025)Sahoo, Deschenaux, Gokaslan, Wang, Chiu, and Kuleshov]{sahoo_diffusion_2025}
Subham~Sekhar Sahoo, Justin Deschenaux, Aaron Gokaslan, Guanghan Wang, Justin Chiu, and Volodymyr Kuleshov.
\newblock The {Diffusion} {Duality}.
\newblock In \emph{Forty-second {International} {Conference} on {Machine} {Learning} ({ICML})}, 2025.
\newblock URL \url{https://openreview.net/forum?id=9P9Y8FOSO}.

\bibitem[Shariatian et~al.(2025)Shariatian, Durmus, and Peluchetti]{shariatian_latent_2025}
Dario Shariatian, Alain Durmus, and Stefano Peluchetti.
\newblock Latent {Discrete} {Diffusion} {Models}.
\newblock \emph{arXiv preprint arXiv:2510.18114}, October 2025.
\newblock \doi{10.48550/arXiv.2510.18114}.
\newblock URL \url{http://arxiv.org/abs/2510.18114}.

\bibitem[Shi et~al.(2024)Shi, Han, Wang, Doucet, and Titsias]{shi_simplified_2024}
Jiaxin Shi, Kehang Han, Zhe Wang, Arnaud Doucet, and Michalis~K Titsias.
\newblock Simplified and {Generalized} {Masked} {Diffusion} for {Discrete} {Data}.
\newblock In \emph{Advances in {Neural} {Information} {Processing} {Systems} ({NeurIPS})}, 2024.
\newblock URL \url{https://openreview.net/forum?id=xcqSOfHt4g}.

\bibitem[Sohl-Dickstein et~al.(2015)Sohl-Dickstein, Weiss, Maheswaranathan, and Ganguli]{sohl-dickstein_deep_2015}
Jascha Sohl-Dickstein, Eric Weiss, Niru Maheswaranathan, and Surya Ganguli.
\newblock Deep {Unsupervised} {Learning} using {Nonequilibrium} {Thermodynamics}.
\newblock In \emph{Proceedings of the 32nd {International} {Conference} on {Machine} {Learning} ({ICML})}, volume~32, June 2015.
\newblock URL \url{https://proceedings.mlr.press/v37/sohl-dickstein15.html}.

\bibitem[Song et~al.(2020)Song, Sohl-Dickstein, Kingma, Kumar, Ermon, and Poole]{song_scorebased_2020}
Yang Song, Jascha Sohl-Dickstein, Diederik~P. Kingma, Abhishek Kumar, Stefano Ermon, and Ben Poole.
\newblock Score-{Based} {Generative} {Modeling} through {Stochastic} {Differential} {Equations}.
\newblock In \emph{International {Conference} on {Learning} {Representations} ({ICLR})}, October 2020.
\newblock URL \url{https://openreview.net/forum?id=PxTIG12RRHS}.

\bibitem[Strudel et~al.(2022)Strudel, Tallec, Altché, Du, Ganin, Mensch, Grathwohl, Savinov, Dieleman, Sifre, and Leblond]{strudel_selfconditioned_2022}
Robin Strudel, Corentin Tallec, Florent Altché, Yilun Du, Yaroslav Ganin, Arthur Mensch, Will Grathwohl, Nikolay Savinov, Sander Dieleman, Laurent Sifre, and Rémi Leblond.
\newblock Self-conditioned {Embedding} {Diffusion} for {Text} {Generation}.
\newblock \emph{arXiv preprint arXiv:211.04236}, November 2022.
\newblock \doi{10.48550/arXiv.2211.04236}.
\newblock URL \url{http://arxiv.org/abs/2211.04236}.

\bibitem[Tack et~al.(2026)Tack, Lanchantin, Yu, Cohen, Kulikov, Lan, Hao, Tian, Weston, and Li]{tack_llm_2026}
Jihoon Tack, Jack Lanchantin, Jane Yu, Andrew Cohen, Ilia Kulikov, Janice Lan, Shibo Hao, Yuandong Tian, Jason Weston, and Xian Li.
\newblock {LLM} {Pretraining} with {Continuous} {Concepts}.
\newblock In \emph{The {Fourteenth} {International} {Conference} on {Learning} {Representations} ({ICLR})}, 2026.
\newblock URL \url{https://openreview.net/forum?id=wTGcb3DxOn}.

\bibitem[Tang et~al.(2025)Tang, Wu, Yang, Xie, Chen, Chen, Zhang, Cai, Lu, and Han]{tang_hart_2025}
Haotian Tang, Yecheng Wu, Shang Yang, Enze Xie, Junsong Chen, Junyu Chen, Zhuoyang Zhang, Han Cai, Yao Lu, and Song Han.
\newblock {HART}: {Efficient} {Visual} {Generation} with {Hybrid} {Autoregressive} {Transformer}.
\newblock In \emph{The {Thirteenth} {International} {Conference} on {Learning} {Representations} ({ICLR})}, 2025.
\newblock URL \url{https://openreview.net/forum?id=q5sOv4xQe4}.

\bibitem[Vaswani et~al.(2017)Vaswani, Shazeer, Parmar, Uszkoreit, Jones, Gomez, Kaiser, and Polosukhin]{vaswani_attention_2017}
Ashish Vaswani, Noam Shazeer, Niki Parmar, Jakob Uszkoreit, Llion Jones, Aidan~N. Gomez, Lukasz Kaiser, and Illia Polosukhin.
\newblock Attention {Is} {All} {You} {Need}.
\newblock In \emph{Advances in {Neural} {Information} {Processing} {Systems} ({NeurIPS})}, volume~30, 2017.
\newblock URL \url{https://proceedings.neurips.cc/paper_files/paper/2017/file/3f5ee243547dee91fbd053c1c4a845aa-Paper.pdf}.

\bibitem[Wang et~al.(2025)Wang, Schiff, Sahoo, and Kuleshov]{wang_remasking_2025}
Guanghan Wang, Yair Schiff, Subham~Sekhar Sahoo, and Volodymyr Kuleshov.
\newblock Remasking {Discrete} {Diffusion} {Models} with {Inference}-{Time} {Scaling}.
\newblock In \emph{The {Thirty}-ninth {Annual} {Conference} on {Neural} {Information} {Processing} {Systems} ({NeurIPS})}, 2025.
\newblock URL \url{https://openreview.net/forum?id=IJryQAOy0p}.

\bibitem[Wei et~al.(2022)Wei, Wang, Schuurmans, Bosma, Ichter, Xia, Chi, Le, and Zhou]{wei_chainofthought_2022}
Jason Wei, Xuezhi Wang, Dale Schuurmans, Maarten Bosma, Brian Ichter, Fei Xia, Ed~H. Chi, Quoc~V. Le, and Denny Zhou.
\newblock Chain-of-{Thought} {Prompting} {Elicits} {Reasoning} in {Large} {Language} {Models}.
\newblock In \emph{Advances in {Neural} {Information} {Processing} {Systems} ({NeurIPS})}, volume~36, October 2022.
\newblock URL \url{https://openreview.net/forum?id=_VjQlMeSB_J}.

\bibitem[Wei et~al.(2025)Wei, Zhang, Liu, Liu, and Zhang]{wei_accelerating_2025}
Qingyan Wei, Yaojie Zhang, Zhiyuan Liu, Dongrui Liu, and Linfeng Zhang.
\newblock Accelerating {Diffusion} {Large} {Language} {Models} with {SlowFast} {Sampling}: {The} {Three} {Golden} {Principles}.
\newblock \emph{arXiv preprint arXiv:2506.10848}, June 2025.
\newblock \doi{10.48550/arXiv.2506.10848}.
\newblock URL \url{http://arxiv.org/abs/2506.10848}.

\bibitem[Wu et~al.(2026)Wu, Zhang, Xue, Liu, Diao, Zhu, Luo, Han, and Xie]{wu_fastdllm_2026}
Chengyue Wu, Hao Zhang, Shuchen Xue, Zhijian Liu, Shizhe Diao, Ligeng Zhu, Ping Luo, Song Han, and Enze Xie.
\newblock Fast-{dLLM}: {Training}-free {Acceleration} of {Diffusion} {LLM} by {Enabling} {KV} {Cache} and {Parallel} {Decoding}.
\newblock In \emph{The {Fourteenth} {International} {Conference} on {Learning} {Representations} ({ICLR})}, 2026.
\newblock URL \url{https://openreview.net/forum?id=3Z3Is6hnOT}.

\bibitem[Wu et~al.(2025)Wu, Teng, and Tu]{wu_parallel_2025}
Haoyi Wu, Zhihao Teng, and Kewei Tu.
\newblock Parallel {Continuous} {Chain}-of-{Thought} with {Jacobi} {Iteration}.
\newblock In \emph{Proceedings of the 2025 {Conference} on {Empirical} {Methods} in {Natural} {Language} {Processing} ({EMNLP})}, pp.\  914--926, 2025.
\newblock \doi{10.18653/v1/2025.emnlp-main.47}.
\newblock URL \url{https://aclanthology.org/2025.emnlp-main.47/}.

\bibitem[Xie et~al.(2025)Xie, Ye, Zheng, Gao, Dong, Wu, Zhao, Gong, Jiang, Li, and Kong]{xie_dreamcoder_2025}
Zhihui Xie, Jiacheng Ye, Lin Zheng, Jiahui Gao, Jingwei Dong, Zirui Wu, Xueliang Zhao, Shansan Gong, Xin Jiang, Zhenguo Li, and Lingpeng Kong.
\newblock Dream-{Coder} {7B}: {An} {Open} {Diffusion} {Language} {Model} for {Code}.
\newblock \emph{arXiv preprint arXiv:2509.01142}, September 2025.
\newblock \doi{10.48550/arXiv.2509.01142}.
\newblock URL \url{http://arxiv.org/abs/2509.01142}.

\bibitem[Ye et~al.(2024)Ye, Gong, Chen, Zheng, Gao, Shi, Wu, Jiang, Li, Bi, and Kong]{ye_diffusion_2024}
Jiacheng Ye, Shansan Gong, Liheng Chen, Lin Zheng, Jiahui Gao, Han Shi, Chuan Wu, Xin Jiang, Zhenguo Li, Wei Bi, and Lingpeng Kong.
\newblock Diffusion of {Thought}: {Chain}-of-{Thought} {Reasoning} in {Diffusion} {Language} {Models}.
\newblock \emph{Advances in Neural Information Processing Systems (NeurIPS)}, 37, December 2024.
\newblock URL \url{https://proceedings.neurips.cc/paper_files/paper/2024/hash/be30024e7fa2c29cac7a6dafcbb8571f-Abstract-Conference.html}.

\bibitem[Ye et~al.(2025)Ye, Xie, Zheng, Gao, Wu, Jiang, Li, and Kong]{ye_dream_2025}
Jiacheng Ye, Zhihui Xie, Lin Zheng, Jiahui Gao, Zirui Wu, Xin Jiang, Zhenguo Li, and Lingpeng Kong.
\newblock Dream {7B}: {Diffusion} {Large} {Language} {Models}.
\newblock \emph{arXiv preprint arXiv:2508.15487}, August 2025.
\newblock \doi{10.48550/arXiv.2508.15487}.
\newblock URL \url{http://arxiv.org/abs/2508.15487}.

\bibitem[Zhang et~al.()Zhang, He, Yan, Shen, Zhao, Wang, Shen, and Wang]{zhang_soft_02}
Zhen Zhang, Xuehai He, Weixiang Yan, Ao~Shen, Chenyang Zhao, Shuohang Wang, Yelong Shen, and Xin~Eric Wang.
\newblock Soft {Thinking}: {Unlocking} the {Reasoning} {Potential} of {LLMs} in {Continuous} {Concept} {Space}.
\newblock In \emph{The {Thirty}-ninth {Annual} {Conference} on {Neural} {Information} {Processing} {Systems} ({NeurIPS})}, February .
\newblock URL \url{https://openreview.net/forum?id=ByQdHPGKgU}.

\bibitem[Zhao et~al.(2025)Zhao, Gupta, Zheng, and Grover]{zhao_d1_2025}
Siyan Zhao, Devaansh Gupta, Qinqing Zheng, and Aditya Grover.
\newblock d1: {Scaling} {Reasoning} in {Diffusion} {Large} {Language} {Models} via {Reinforcement} {Learning}.
\newblock In \emph{The {Thirty}-ninth {Annual} {Conference} on {Neural} {Information} {Processing} {Systems} ({NeurIPS})}, 2025.
\newblock URL \url{https://openreview.net/forum?id=7ZVRlBFuEv}.

\bibitem[Zheng et~al.(2026)Zheng, Gong, Zhang, Chen, Gu, Zhou, Jaitly, and Zhang]{zheng_continuously_2026}
Huangjie Zheng, Shansan Gong, Ruixiang Zhang, Tianrong Chen, Jiatao Gu, Mingyuan Zhou, Navdeep Jaitly, and Yizhe Zhang.
\newblock Continuously {Augmented} {Discrete} {Diffusion} model for {Categorical} {Generative} {Modeling}.
\newblock In \emph{The {Fourteenth} {International} {Conference} on {Learning} {Representations} ({ICLR})}, 2026.
\newblock URL \url{https://openreview.net/forum?id=JNAZ3e7Bwt}.

\bibitem[Zhou et~al.(2025)Zhou, Yang, Hu, Wang, Zhang, Zhang, Mackey, Jaakkola, Bates, and Zhang]{zhou_coevolutionary_2025}
Cai Zhou, Chenxiao Yang, Yi~Hu, Chenyu Wang, Chubin Zhang, Muhan Zhang, Lester Mackey, Tommi Jaakkola, Stephen Bates, and Dinghuai Zhang.
\newblock Coevolutionary {Continuous} {Discrete} {Diffusion}: {Make} {Your} {Diffusion} {Language} {Model} a {Latent} {Reasoner}.
\newblock \emph{arXiv preprint arXiv:2510.03206}, October 2025.
\newblock \doi{10.48550/arXiv.2510.03206}.
\newblock URL \url{http://arxiv.org/abs/2510.03206}.

\bibitem[Zhu et~al.(2025{\natexlab{a}})Zhu, Wang, Nie, Zhang, Wu, Hu, Zhou, Chen, Lin, Wen, and Li]{zhu_llada_2025}
Fengqi Zhu, Rongzhen Wang, Shen Nie, Xiaolu Zhang, Chunwei Wu, Jun Hu, Jun Zhou, Jianfei Chen, Yankai Lin, Ji-Rong Wen, and Chongxuan Li.
\newblock {LLaDA} 1.5: {Variance}-{Reduced} {Preference} {Optimization} for {Large} {Language} {Diffusion} {Models}.
\newblock \emph{arXiv preprint arXiv:2505.19223}, May 2025{\natexlab{a}}.
\newblock \doi{10.48550/arXiv.2505.19223}.
\newblock URL \url{http://arxiv.org/abs/2505.19223}.

\bibitem[Zhu et~al.(2025{\natexlab{b}})Zhu, Hao, Hu, Jiao, Russell, and Tian]{zhu_reasoning_2025}
Hanlin Zhu, Shibo Hao, Zhiting Hu, Jiantao Jiao, Stuart Russell, and Yuandong Tian.
\newblock Reasoning by {Superposition}: {A} {Theoretical} {Perspective} on {Chain} of {Continuous} {Thought}.
\newblock In \emph{The {Thirty}-ninth {Annual} {Conference} on {Neural} {Information} {Processing} {Systems} ({NeurIPS})}, 2025{\natexlab{b}}.
\newblock URL \url{https://openreview.net/forum?id=UdOEZgWJLc}.

\bibitem[Zhuang et~al.(2025)Zhuang, Liu, Singh, Shang, and Gao]{zhuang_mixture_2025}
Yufan Zhuang, Liyuan Liu, Chandan Singh, Jingbo Shang, and Jianfeng Gao.
\newblock Mixture of {Inputs}: {Text} {Generation} {Beyond} {Discrete} {Token} {Sampling}.
\newblock In \emph{The {Thirty}-ninth {Annual} {Conference} on {Neural} {Information} {Processing} {Systems} ({NeurIPS})}, 2025.
\newblock URL \url{https://openreview.net/forum?id=l6C6Pw30Gl}.

\end{thebibliography}
